\PassOptionsToPackage{pdfversion=1.7}{hyperref}
\documentclass[
  sora,
  strictfonts,
  numberedsections,
  onecolumn
]{mirrosarticle}

\usepackage{xspace}
\usepackage{algorithm}
\usepackage{algpseudocode}
\usetikzlibrary{arrows.meta,calc,fit,matrix,positioning,backgrounds,patterns,decorations.pathreplacing}

\newcommand{\corrauthor}{\textsuperscript{\textdagger}}

\newcommand{\Expect}{\mathop{\mathbf{E}}\nolimits}
\newcommand{\sg}{\operatorname{sg}}

\newcommand{\rgbtokens}{390\xspace}     %
\newcommand{\geotokens}{28\xspace}      %
\newcommand{\blockfps}{35\xspace}       %

\colorlet{FigFill}{MirrosLavender!16}
\colorlet{FigFillStrong}{MirrosLavender!55}
\colorlet{FigStroke}{MirrosPurpleSoft!75}
\colorlet{FigEdge}{MirrosMuted!75}
\colorlet{FigGrey}{MirrosCardBg}
\tikzset{
  fig/.style={font=\sffamily\small, >={Stealth[length=2.2mm]}},
  box/.style={draw=FigStroke, fill=FigFill, rounded corners=3pt, line width=0.6pt,
    align=center, inner sep=5pt, minimum height=9mm},
  greybox/.style={draw=MirrosLine!80!black, fill=FigGrey, rounded corners=3pt,
    line width=0.6pt, align=center, inner sep=5pt, minimum height=9mm},
  pill/.style={draw=FigEdge, dashed, rounded corners=8pt, line width=0.6pt,
    align=center, inner sep=5pt, minimum height=9mm},
  flow/.style={->, draw=FigEdge, line width=0.7pt},
  accentflow/.style={->, draw=MirrosPurple, line width=1pt},
  lbl/.style={font=\sffamily\scriptsize, text=MirrosMuted, align=center},
  heading/.style={font=\sffamily\bfseries\small, text=MirrosInk},
  stepbadge/.style={circle, fill=MirrosPurple, text=white, font=\sffamily\bfseries\scriptsize,
    inner sep=0pt, minimum size=4.2mm},
}

\newfontfamily\captionleadfont[Extension=.otf,ItalicFont=lmromandemi10-oblique]{lmromandemi10-regular}
\newcommand{\captitle}[1]{{\captionleadfont #1}}

\appto{\bibsetup}{\sloppy}

\definecolor{revisionblue}{RGB}{0, 102, 204}

\title{AgentGarten: Code Worlds for Evolving Agents}

\author{\textbf{MirroS Technical Report}}
\pdfauthor{MirroS Technical Report}

\tcbset{mirros/frontpanel/.append style={top=7pt,bottom=7pt}}
\date{October 8, 2026}
\metadata[Blog]{\url{https://mirros.ai/blog/worlds-for-evolving-agents}}
\metadata[Code]{\url{https://github.com/MirroS-Lab/AgentGarten}}
\metadata[Project Page]{\url{https://mirros-lab.github.io/agent-garten}}
\keywords{world models, code worlds, executable environments, neural rendering,
autoregressive video generation, distribution matching distillation, agents}
\runningtitle{AgentGarten: Code Worlds for Evolving Agents}

\begin{abstract}
  Interactive virtual worlds allow agents to learn through
  exploration and interaction. What agents can learn is bounded by the
  environments they practice in, which must be \emph{faithful}, with consistent state,
  rules, and dynamics, and
  \emph{realistic}, with observations that follow the real-world visual
  distributions. Achieving both across diverse worlds remains a bottleneck.
  We introduce \emph{AgentGarten}, a framework that couples simulators and game engines with a shared \emph{neural renderer} to build real-time interactive environments.
  Its simulation backends maintain persistent world state and execute
  program-defined interaction rules, while the renderer
  generates visual observations from structured conditions exported through a common
  interface.
  To build the neural renderer, we adapt a pretrained video model to geometry
  conditions, distill it with our proposed \emph{Adversarial Forcing}, and optimize
  inference for real-time interaction. Adversarial Forcing makes history prefilling
  differentiable through exact replay, so that losses on later predictions
  update how the renderer encodes prior observations, and adds real-data
  adversarial supervision to improve its visual quality.
  In AgentGarten, agents perceive the world through visual observations, interact with it in real time, and improve by distilling each round of experience into playbooks that subsequent agents inherit and refine. Our empirical study demonstrates a substantial gain in learning efficiency, with agents learning from just 4 rounds compared with millions for a conventional reinforcement learning counterpart. As new worlds can be written as code and rendered
  through the same interface, environments can scale in both number and difficulty
  alongside their agents, a step toward agents that keep evolving through
  interactive experience.
\end{abstract}

\teaserfigure[\captitle{Code worlds for evolving agents.} \textbf{Top}: Diverse environments spanning distinct geometries, visual appearances, and interactive tasks. \textbf{Bottom}:
Agents continuously evolve through environmental interactions, developing increasingly sophisticated strategies such as shelter-building and ramp-crossing.\label{fig:teaser}]{%
  \includegraphics[width=0.97\linewidth]{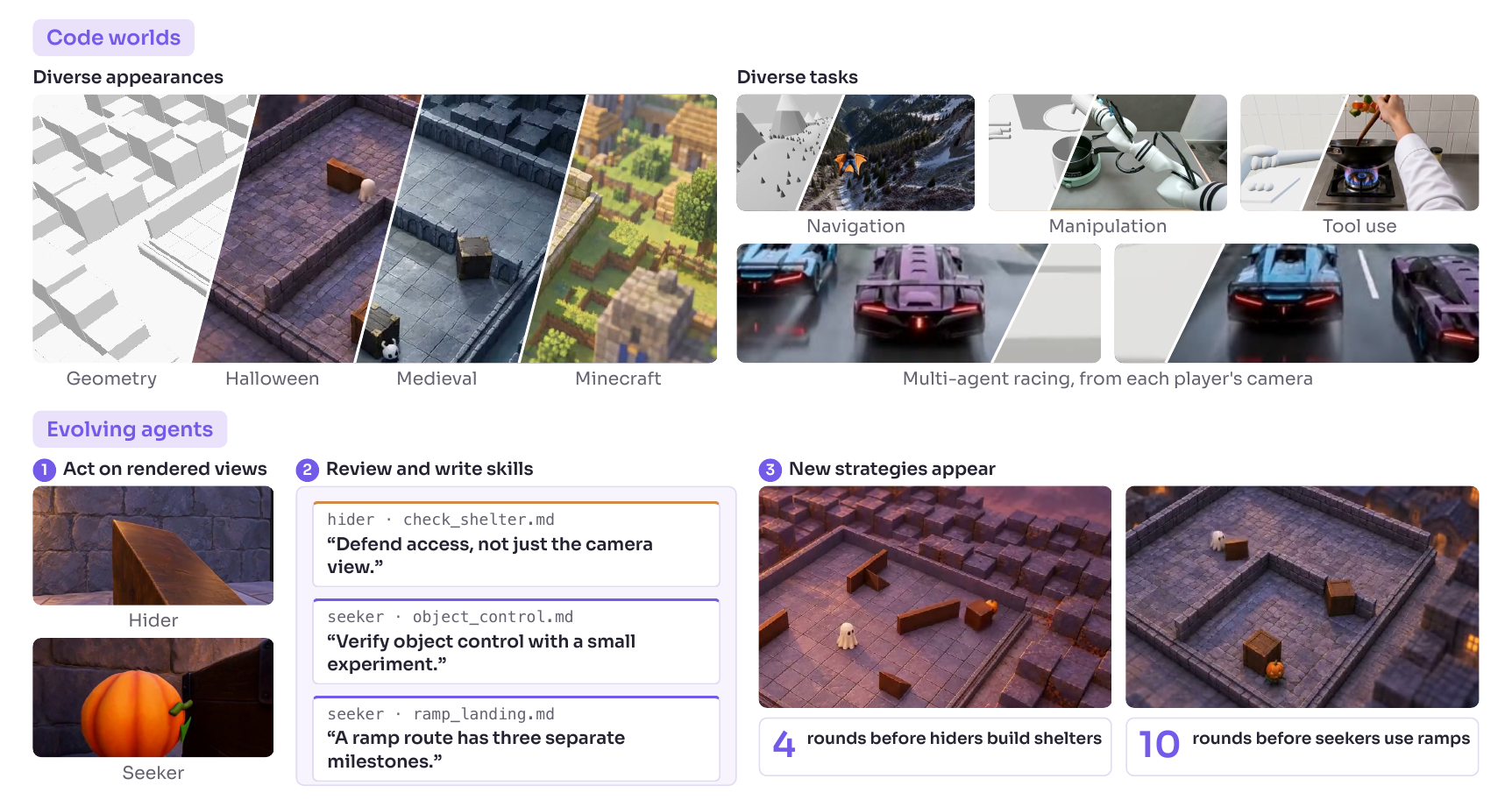}%
}

\begin{document}
\maketitle

\clearpage
\section{Introduction}
\label{sec:introduction}

Agents learn from trajectories of actions and observations. Developing
scalable, generalizable agent capabilities requires environments that not only
faithfully preserve the consequences of past actions over extended
interactions, but also support flexible variation in layouts, objects, and task
rules. Across navigation, manipulation, and multi-agent interaction, scaling
these environments therefore requires both controllable, inspectable dynamics
and diverse visual observations.

However, existing approaches face a fundamental trade-off. Simulators and game
engines support agent learning through explicit state and programmable
interaction rules \citep{szot2021habitat2,gu2023maniskill2,deitke2022procthor}.
Yet expanding their visual diversity entails substantial 3D asset creation and
complex rendering pipelines, making new environments costly to build
and customize.
Conversely, video world models synthesize rich, interactive visual observations
from data
\citep{bruce2024genie,sun2025worldplay,wang2026matrixgame3,gao2026lingbotworld2}.
In these models, however, task-relevant state and physical transition
rules remain
implicit within generated histories and learned representations,
precluding direct
inspection, editing, and testing of the environment's behavior.

\begin{figure}[t]
  \centering
  \includegraphics[width=\linewidth]{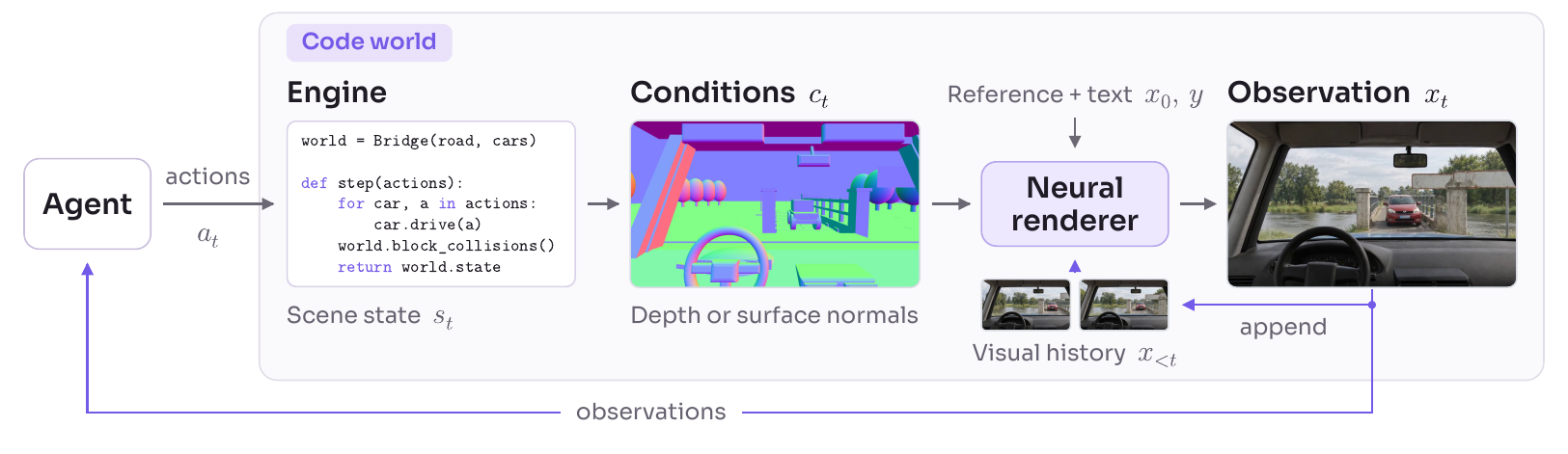}
  \caption{\captitle{Interaction and rendering in a code world}
  (Eqs.~\ref{eq:code-world} and~\ref{eq:renderer}). The agent submits actions
  and receives generated observations. The engine maintains scene
  state and exports structured conditions, here surface normals; the neural
  renderer combines them
  with an appearance reference, text, and cached visual history. New observations
  return to the agent and join the history.}
  \label{fig:loop}
\end{figure}

We introduce \emph{AgentGarten}, a framework for building \emph{code worlds}
that combines programmable dynamics with a shared, real-time neural renderer
(Figure~\ref{fig:loop}). In AgentGarten, each scene program runs in a simulator
or game engine that maintains persistent state and executes explicit interaction
rules. A shared neural renderer synthesizes the agent's visual observations from
structured conditions exported by the engine through a unified interface. This
separation preserves rigorous control over state and rules while
allowing compatible
engines to share a single renderer. New environments can thus be authored by
coding agents from text or image inputs, then edited and extended entirely in
code without creating bespoke visual assets for every scene.

Operating a video model as an interactive neural renderer requires it
to strictly
follow updated engine conditions, maintain visual consistency across extended
rollouts, and synthesize observations in fine-grained, short blocks so that
agents receive immediate feedback after brief actions. We adapt a pretrained
bidirectional video model to structured conditions, convert it to block-causal
generation, and distill it into a real-time renderer on a single GPU with
\emph{Adversarial Forcing}, which trains the renderer on its own rollouts via
distribution matching \citep{huang2025selfforcing,yin2023dmd,yin2024dmd2} and
introduces two key advances. First, to allow later losses to update the
history-encoding computation without the memory cost of a fully differentiable
rollout, we adopt two-pass training \citep{zhuang2026selfgradient} and
introduce \emph{exact replay}: a
block-by-block execution schedule that eliminates numerical divergence between
sampling and recomputation and achieves bitwise-identical recomputation of the
rollout trajectory. Second, to
prevent the visual degradation that score distillation alone exhibits over long
rollouts, we incorporate a real-data adversarial objective
\citep{yin2024dmd2,huang2024r3gan} and derive an exact R1/R2 regularization
scheme for a discriminator with a frozen backbone, entirely avoiding
double-backward passes through fused attention kernels.

Code worlds provide rich environments for agents to practice and evolve. We
evaluate pretrained foundation-model agents across rounds of interaction,
where they perceive the world strictly through rendered observations, diagnose
failures, and distill key insights into written \emph{playbooks} inherited by
subsequent rounds. In hide-and-seek \citep{baker2020emergent}, agents
rapidly grounded abstract spatial knowledge into closed-loop physical
execution, with ramp use and shelter construction emerging within a handful of
games. The same reflective loop improved performance across rounds in several
additional, diverse environments (Section~\ref{sec:agents}).

Our contributions are:
\begin{contributions}
\item A framework for code worlds that couples persistent, editable environment
  state to a shared real-time neural renderer through structured conditions,
  enabling programmable interaction and diverse visual observations
  across compatible
  engines without per-scene visual asset authoring.
\item \emph{Adversarial Forcing}, a distillation method for a few-step,
  block-causal neural renderer. It combines distribution matching on
  self-rollouts
  with exact bitwise replay for history gradients, alongside a real-data
  adversarial objective whose exact R1/R2 regularization avoids double-backward
  through fused attention.
\item An empirical study of evolving agents. Pretrained agents
  observing exclusively
  through neural rendering rapidly ground semantic tool knowledge and improve
  outcomes across rounds through written playbooks in hide-and-seek and several
  further distinct worlds.
\end{contributions}

\section{Code Worlds}
\label{sec:code-world}

\subsection{Formulation}
\label{sec:formulation}

A code world combines a scene program $p$, an engine that executes it, and a
neural renderer $R_\theta$. The program specifies the scene and interaction
rules. Let $s_t$ denote the scene state, including object poses, articulations,
and task variables; $\pi_t$ the observing camera pose; and $a_t$ the actions of
one or more agents. The engine updates the state and captures a structured
condition $c_t$ from the camera:
\begin{equation}
  (s_{t+1},\pi_{t+1}) = f_p(s_t,\pi_t,a_t),
  \qquad
  c_t = h_p(s_t,\pi_t),
  \label{eq:code-world}
\end{equation}
where $f_p$ implements the state transition and $h_p$ renders the visible
geometry. Given an appearance reference $x_0$ and a text description $y$, the
neural renderer generates subsequent observations from visual history $x_{<t}$:
\begin{equation}
  x_t = R_\theta\!\left(x_{<t},\,c_{\le t},\,x_0,\,y\right),
  \qquad t\geq1.
  \label{eq:renderer}
\end{equation}
The agent receives $x_t$ and selects $a_t$, which determines the next state and
condition through Eq.~\eqref{eq:code-world}.

Unlike a video world model, whose state resides in generated history, the
transition $f_p$ takes no rendered observation as input: the renderer
influences the state only through the actions that agents select, so rendering
errors cannot accumulate in the state. A recorded state trajectory can
consequently be rendered again under a different appearance reference or
camera without altering the underlying events. Conversely, the renderer
observes the state only through $c_{\le t}$; attributes that the conditions
leave undetermined, such as color and material, are specified by $x_0$ and $y$
and carried by the visual history. The renderer consumes conditions and visual
history in blocks, as described in Section~\ref{sec:renderer}.

\subsection{Structured conditions}
\label{sec:conditions}

We use colorized depth or surface normals as structured conditions. Both can
be estimated from video or rendered by an engine, and their three-channel
representation allows us to reuse the pretrained video encoder and
Transformer. For real videos, we obtain depth with
ViPE~\citep{huang2025vipe}, using Depth Anything~3~\citep{lin2025depthanything3}
as its depth backend, and normals with NormalCrafter~\citep{bin2025normalcrafter}.
Simulated scenes supply these maps directly from geometry, without requiring
detailed materials or textures. During training, each sample uses one randomly
chosen modality (Section~\ref{sec:geometry-conditioning}).

We encode depth with the Vision Banana mapping~\citep{visionbanana2026} and
camera-space unit normals as $(n+1)/2$, using consistent coordinate and
orientation conventions across estimated and rendered maps
(Appendix~\ref{app:geometry}).

\subsection{Building code worlds}
\label{sec:building}

A coding agent constructs a code world from either a single image or a text description, producing an executable scene program along with an appearance reference $x_0$. Because appearance is delegated to the neural renderer, the scene program does not require detailed materials or leaf-level assets; it only needs coarse geometry that faithfully conveys layout, silhouettes, occlusion, and motion dynamics.

\begin{figure}[t]
  \centering
  \includegraphics[width=\linewidth]{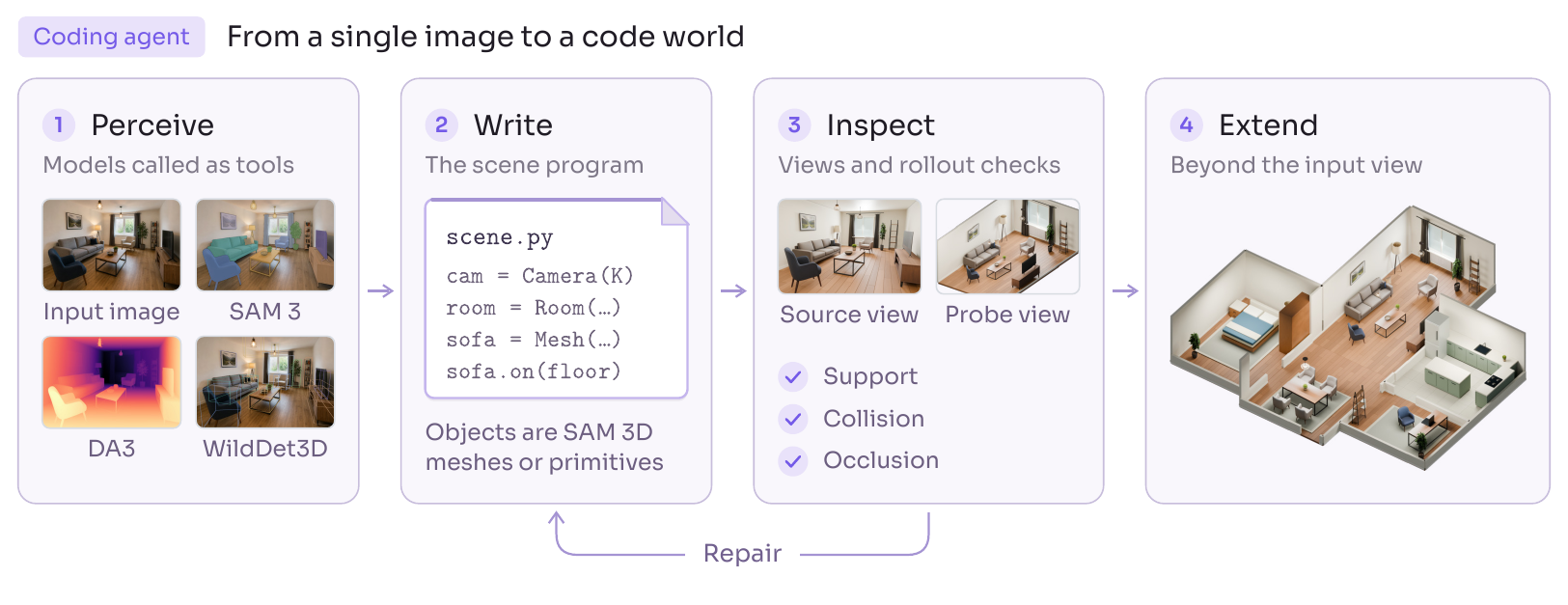}
  \caption{\captitle{A coding agent builds a code world from a single image.}
    A living-room example. The agent calls perception and generation models as
    tools, writes a scene program from their outputs, and repairs it after
    inspecting rendered views and rollout checks. The expanded scene keeps the
    observed room and adds connected rooms beyond the input view as authored
    extensions.
  The input image supplies the appearance reference $x_0$.}
  \label{fig:image-to-3d}
\end{figure}

\paragraph{From an image.}
The input image serves simultaneously as $x_0$ and the reference layout (Figure~\ref{fig:image-to-3d}). Perception and generation models lift the visible content into instance masks \citep{carion2025sam3}, monocular depth and camera intrinsics \citep{lin2025depthanything3}, 3D bounding boxes \citep{wilddet3d}, and extracted object meshes \citep{sam3dteam2025sam3d3dfyimages}. A coding agent fits these assets to the estimated geometry, completes unobserved regions as plausible spatial extensions, and binds physical properties and interaction rules.

\paragraph{From text and agentic refinement.}
When building from text, the agent writes the scene program directly using geometric primitives and engine assets. An initial render is transformed into $x_0$ via an image editing model while preserving scene layout. In both modalities, the agent refines the scene through an interactive feedback loop: it evaluates test rollouts from multiple probe cameras, inspects engine verification queries (contact, collision clearance, and occlusion), and iteratively repairs faulty poses, unsupported structures, or ambiguous motions before deployment.

\section{A Real-Time Neural Renderer}
\label{sec:renderer}

We train a geometry-conditioned autoregressive video model to implement the
renderer in Section~\ref{sec:formulation}. Starting from a pretrained
bidirectional backbone, training proceeds through geometry conditioning,
teacher-forcing adaptation, and \emph{Adversarial Forcing}. The final stage
combines distribution matching on the model's own rollouts with exact
history-gradient replay and real-data adversarial training. A bounded
key--value cache and streaming decoding
support continuous inference within the interaction loop in
Figure~\ref{fig:loop}.

\subsection{Backbone}
\label{sec:backbone}

We initialize from Cosmos~3-Nano \citep{nvidia2026cosmos3} and separate its
understanding (UND) and generation (GEN) towers. The frozen UND tower encodes
the caption $y$ into per-layer keys and values consumed by GEN. This separation
allows independent compilation and distributed execution of the two towers;
during distillation, the student, teacher, and fake-score networks share one
UND tower and reuse its outputs for the same caption.

A frozen Wan video autoencoder \citep{wanteam2025wan} maps RGB and geometry
videos to latents. The appearance reference $x_0$ uses the backbone's native
clean-frame representation: it receives no diffusion timestep and is never
denoised. Architecture sizes and training configurations are given in
Appendix~\ref{app:stages}.

\subsection{Geometry conditioning}
\label{sec:geometry-conditioning}

\paragraph{Condition tokens.}
Depth and normals use the three-channel encodings of
Section~\ref{sec:conditions}. Spatial average pooling reduces the condition
encoder's input size and the number of tokens processed during training and
inference. For an encoded condition frame indexed by $t$, we form
\begin{equation}
  G_t = W_{\mathrm{in}}\,\mathcal{P}\!\big([E(S(c))]_t\big) + e_{\mathrm{geo}},
  \label{eq:condition-tokens}
\end{equation}
where $S$ pools the condition video spatially, $E$ is the frozen video encoder,
and $\mathcal{P}$ groups latent patches. We reuse the pretrained RGB input
projection $W_{\mathrm{in}}$ and add a zero-initialized modality embedding
$e_{\mathrm{geo}}$. Condition tokens receive no diffusion timestep.

\paragraph{Joint attention.}
We concatenate geometry and RGB tokens along the sequence dimension and
process them with the same Transformer layers. We use neither channel
concatenation
\citep{nvidia2026omnidreams} nor control branches
\citep{zhang2023controlnet,nvidia2025cosmos25}, to avoid imposing a
pixel-aligned inductive bias. Writing $R_t$ for RGB latent tokens, the visual
sequence is
\begin{equation}
  [\,G_0,G_1,\dots\,]\;[\,R_0,R_1,\dots\,],
\end{equation}
with $R_0$ encoding $x_0$ and text keys and values integrated via
cross-attention. Geometry
and RGB share temporal rotary coordinates; spatial coordinates map both grids
to the same image extent. This preserves geometric alignment while allowing
cross-token interactions. Only noisy RGB tokens produce velocity predictions.

\paragraph{Robust conditions.}
Geometry estimated from natural video differs from engine-rendered geometry
in texture leakage, alignment, and missing observations. To reduce dependence
on estimator-specific cues, we randomly retain depth or normals, with
occasional dropout of both modalities; dropped inputs are black. We suppress
fine texture by down- and upsampling, perturb alignment with a smooth spatial
warp anchored to the first frame, and add noise to condition latents.
Preprocessing and augmentation settings appear in Appendix~\ref{app:geometry}.

\subsection{Teacher-forcing adaptation}
\label{sec:causal}

Stage 2 adapts the geometry-conditioned model to blockwise autoregressive
generation. We partition the latents after the reference into fixed-length
blocks. Each block is generated from its aligned geometry and previously
completed blocks, without access to future blocks.

\begin{figure}[t]
  \centering
  \begin{minipage}[c]{0.56\linewidth}
    \includegraphics[width=\linewidth]{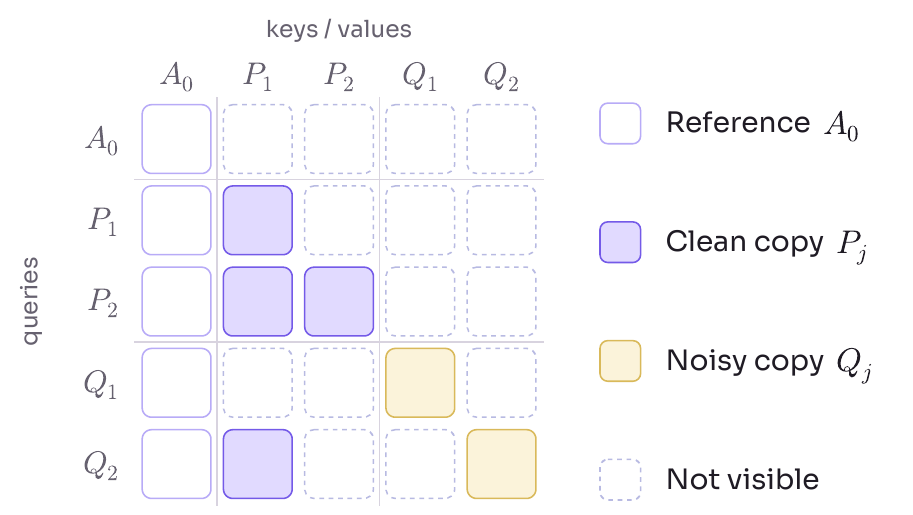}
  \end{minipage}\hfill
  \begin{minipage}[c]{0.40\linewidth}
  \caption{\captitle{Block-causal teacher-forcing mask, shown for two target
  blocks.} Clean
  copies $P_j$ encode history and see the reference, earlier clean blocks, and
  themselves;
  noisy copies $Q_j$ see the reference, clean blocks before $j$, and themselves,
  but never their own clean target. Text is visible to every row. The same
  layout is reused, with recorded latents, by the replay pass of Adversarial
  Forcing.}
  \label{fig:ar-mask}
  \end{minipage}
\end{figure}

\paragraph{Parallel teacher forcing.}
Training uses two copies of each block inside one Transformer: a clean copy
$P_j$ that represents history and a noisy copy $Q_j$ that is denoised. Each
copy carries its own aligned condition tokens. With $A_0$ the reference, the
attention mask permits
\begin{equation}
  P_j \longrightarrow A_0\cup P_{\le j},
  \qquad
  Q_j \longrightarrow A_0\cup P_{<j}\cup Q_j,
  \label{eq:ar-attention-mask}
\end{equation}
where an arrow points from a query to the tokens it may read; text is visible
to both (Figure~\ref{fig:ar-mask}). The noisy copy of a block therefore sees
clean earlier blocks and attends bidirectionally within itself, but never its
own clean target. This lets all target blocks be trained in parallel with
independently sampled flow times. With $z_j$ the clean latent block and
$\mathcal C=(x_0,y,c)$ the reference, text, and conditions,
\begin{equation}
  \begin{aligned}
    z_{j,t_j} &= (1-t_j)\,z_j+t_j\,\epsilon_j,
    \qquad \epsilon_j\sim\mathcal{N}(0,I),\\
    \mathcal{L}_{\mathrm{TF}}
    &= \Expect\!\left[\frac{1}{M}\sum_{j=1}^{M}
      \big\|v_\theta(z_{j,t_j},t_j\mid z_{<j},\mathcal{C})
    -(\epsilon_j-z_j)\big\|_2^2\right],
  \end{aligned}
  \label{eq:ar-flow-training}
\end{equation}
where $M$ is the number of target blocks and conditioning on $z_{<j}$ is
implemented by the clean copies and the mask.

\subsection{Adversarial Forcing}
\label{sec:self-forcing}
\label{sec:adversarial-forcing}

Stage 3, \emph{Adversarial Forcing}, distills the teacher-forced model into a
few-step renderer that is trained on its own rollouts. It has three
components: distribution matching against the bidirectional teacher, exact
replay of each rollout so that gradients reach the history it wrote, and a
real-data adversarial objective with exact R1/R2 regularization.

\subsubsection{Distribution matching}
\label{sec:distribution-matching}

We match autoregressive student rollouts to the bidirectional teacher through
distribution matching distillation (DMD) in the manner of Self Forcing
\citep{huang2025selfforcing,yin2023dmd,yin2024dmd2}. The student starts from
stage 2 and generates each block in a few denoising steps, conditioning on its
own preceding outputs. The frozen stage-1 teacher scores the resulting clip
jointly. An auxiliary \emph{fake-score} model, initialized from the teacher,
learns the student's distribution.

For a student prediction $\widetilde z_\theta$ corrupted to
$y_\tau=(1-\tau)\widetilde z_\theta+\tau\epsilon$, let $f_{\mathrm T}$ and
$f_\psi$ denote clean estimates from the teacher and fake-score model. The
student objective is
\begin{equation}
  \begin{aligned}
    g &= \frac{f_\psi(y_\tau,\tau\mid\mathcal{C})-
    f_{\mathrm{T}}(y_\tau,\tau\mid\mathcal{C})}{a},
    \qquad
    \mathcal{L}_{\mathrm{DMD}}
    = \Expect\!\left[\frac{1}{2N}
      \big\|\widetilde z_\theta-
    \sg(\widetilde z_\theta-g)\big\|_2^2\right],
  \end{aligned}
  \label{eq:dmd}
\end{equation}
where $a$ is a per-video normalization, $N$ the number of latent elements, and
$\sg$ stops gradients. The fake-score model is trained by flow matching on
detached student samples. Sampling schedules, guidance, and update ratios are
specified in Appendix~\ref{app:distillation}.

The other two components extend this objective: replay restores gradients
through history encoding, and an adversarial loss supplies supervision from
real videos.

\subsubsection{Exact replay}
\label{sec:exact-replay}

In standard Self Forcing, completed blocks are encoded into a detached KV
cache. Subsequent losses therefore supervise prediction from that cache, but
not the history-prefill computation that produced its keys and values.
Keeping this computation differentiable in a single serial rollout would
retain the recursively connected cache-formation graphs across blocks, with
substantial memory cost.

Similar to Self Gradient Forcing (SGF)
\citep{zhuang2026selfgradient}, we decouple rollout and gradient propagation
using two passes. A no-gradient rollout records each block's input
$U$ at the final
denoising time $t^*$ and its clean output $Z$. A differentiable replay then
recomputes both history and predictions with the visibility pattern of
Eq.~\eqref{eq:ar-attention-mask}. For block $j$,
\begin{equation}
  \widetilde z_{\theta,j}
  = \sg(U_j)-t^*\,v_\theta\!\left(
    \sg(U_j),t^*\mid
  \sg(Z_{<j}),\mathcal{C};\mathcal{M}_{\mathrm{TF}}\right).
  \label{eq:sgf-prediction}
\end{equation}
The recorded latents stay detached, but their history encodings are
recomputed inside the graph. Later losses thus update the parameters that
write history into the cache, without differentiating through the sampling
trajectory. Both DMD and generator adversarial losses use this replayed
prediction.

\begin{figure}[t]
  \centering
  \includegraphics[width=\linewidth]{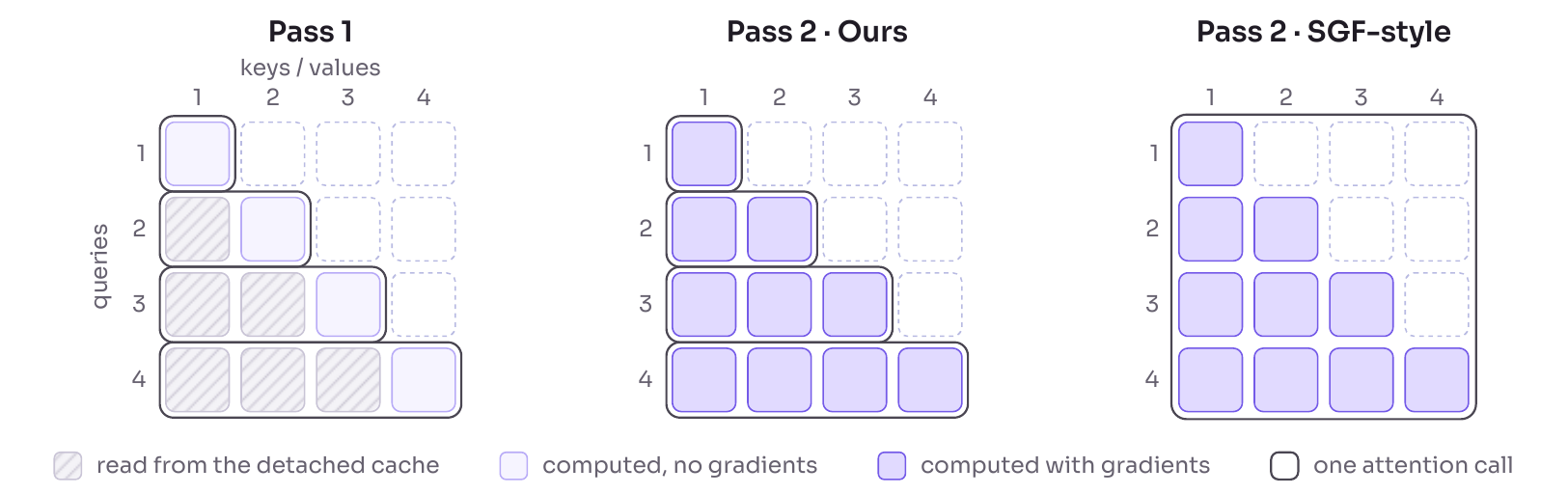}
  \caption{\captitle{Exact replay, drawn as block-level attention masks.} Row
  $i$ holds
  block $i$'s queries, and the columns are the key/value blocks it reads. The
  rollout runs one attention call per block and reads earlier blocks from a
  detached cache. Our replay runs the same calls but computes the history keys
  and values with gradients, preserving the rollout's execution structure.
  An SGF-style replay computes the same mask in one full-sequence call, which
  changes numerical execution. Table~\ref{tab:replay} evaluates the difference.
  In the implementation each
  block issues two calls, one for its noisy queries and one for the clean
  publication that later blocks read.}
  \label{fig:exact-replay}
\end{figure}

SGF implements its second pass with full-sequence FlexAttention
\citep{guessous2024flexattention}.
Although the
attention mask matches cached generation, the kernel, tensor shapes, and
reduction order differ; in finite precision, replay can consequently diverge
from the sampled trajectory. We preserve the rollout's execution structure:
attention runs block by block with scaled dot-product attention (SDPA), using
the same key--value order and call shapes. Projections and feed-forward layers
use the same block grouping as well. The history keys and values are
recomputed and assembled differentiably, so the matching execution retains
the history-gradient path (Figure~\ref{fig:exact-replay}).
Section~\ref{sec:replay-results} evaluates numerical agreement and cost.

\subsubsection{Adversarial training}
\label{sec:gan}

Distribution matching uses teacher and fake-score estimates on generated
samples, without directly contrasting them with real videos. Following DMD2
\citep{yin2024dmd2}, we add an adversarial objective. A trainable head
$h_\phi$ aggregates intermediate features from the frozen teacher backbone
$B$ into a discriminator logit. Conditioning the backbone on geometry and the
reference lets the discriminator assess appearance together with geometric
consistency. Section~\ref{sec:qualitative} compares the resulting renderer
with one trained by Self Forcing alone.

\paragraph{Relativistic objective.}
We use the relativistic pairing of R3GAN \citep{huang2024r3gan}. With $r$ and
$f$ the logits of a real clip and of the student's prediction, corrupted with
the same noise and timestep under the same conditions,
\begin{equation}
  \mathcal{L}_{\mathrm{G}}
  = \Expect\!\left[\operatorname{softplus}(\sg(r)-f)\right],
  \qquad
  \mathcal{L}_{\mathrm{rel}}
  = \Expect\!\left[\operatorname{softplus}(f-r)\right],
  \label{eq:relativistic}
\end{equation}
and the student minimizes $\mathcal L_{\mathrm{DMD}}+\lambda_{\mathrm G}\mathcal
L_{\mathrm G}$. The generator-side gradient is injected into the second-pass
graph through a linear surrogate, so both
objectives act on Eq.~\eqref{eq:sgf-prediction}.

\paragraph{Exact R1/R2 without double backward.}
R3GAN regularizes the discriminator with R1 and R2, the squared input-gradient
norms on real and generated samples:
\begin{equation}
  D_\phi(x)=h_\phi(B(x)),\qquad
  R_1=\Expect_{x\sim p_{\mathrm{data}}}\|\nabla_x D_\phi(x)\|_2^2,\qquad
  R_2=\Expect_{x\sim p_\theta}\|\nabla_x D_\phi(x)\|_2^2 .
\end{equation}
Their parameter gradients normally require differentiating through a backward
pass. The fused attention kernels in our backbone, such as FlashAttention
\citep{dao2022flashattention}, do not support this double backward. APT
\citep{lin2025apt} sidesteps the problem with a random-perturbation
approximation of R1. We compute the exact penalty and its exact
head-parameter gradient instead, using the fact that the backbone is frozen.

\begin{figure}[t]
  \centering
  \includegraphics[width=\linewidth]{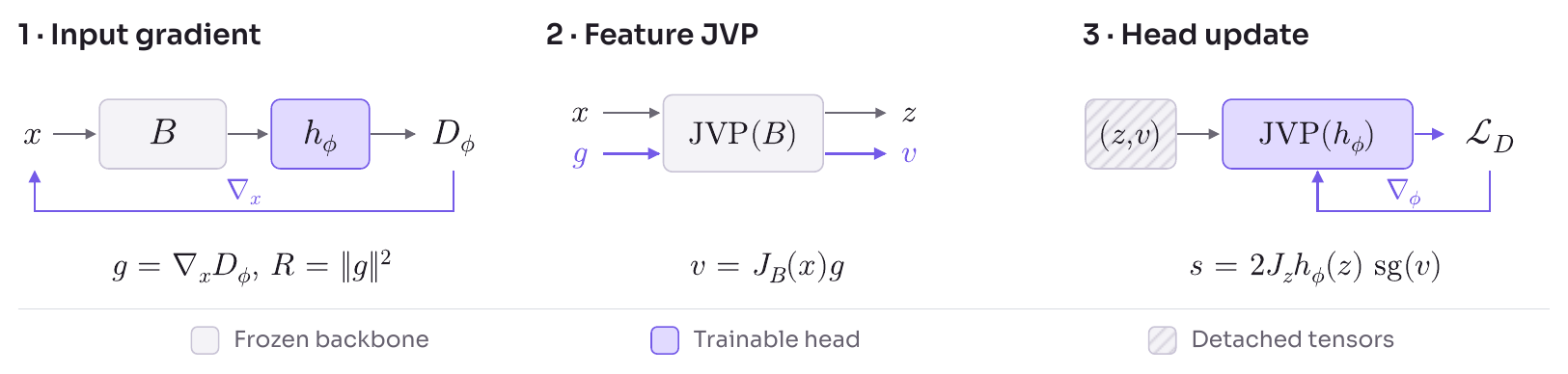}
  \caption{\captitle{Exact R1/R2 with a frozen backbone $B$ and a trainable
  head $h_\phi$.} (1) Differentiate with respect to $x$ only to obtain $g$ and $R$.
  (2) Evaluate the backbone JVP to obtain features $z$ and directions $v$.
  (3) Evaluate the explicit head JVP on detached $(z,v)$, returning both the
  logit and its directional derivative, which supplies $s$. An ordinary
  backward pass with respect to $\phi$ updates the head using the full
  discriminator objective.
  }
  \label{fig:exact-r1r2}
\end{figure}

Consider one sample and write $z=B(x)$, $A=J_B(x)$ and $u=\nabla_z h_\phi(z)$.
Then
\begin{equation}
  g = A^{\top}u,\qquad R=g^{\top}g,\qquad v = A g,
  \label{eq:r1-parts}
\end{equation}
where $g$ is obtained by one vector--Jacobian product (a backward pass through
the head, with its parameters frozen, and then through the backbone) and $v$ by
one Jacobian--vector product that carries $g$ forward through the frozen
backbone without a gradient graph. Neither step materializes $A$. Because $z$
and $A$ do not depend on $\phi$,
\begin{equation}
  \nabla_\phi R
  = 2\left(\frac{\partial u}{\partial \phi}\right)^{\!\top} v
  = \nabla_\phi\!\left[\,2\,J_z h_\phi(z)\,\sg(v)\,\right].
  \label{eq:r1-identity}
\end{equation}
The right-hand side is the gradient of a directional derivative of the head
alone, which we compute by writing the head's Jacobian--vector product with
ordinary differentiable operations (Appendix~\ref{app:r1r2}). With
$s=2J_zh_\phi(z)\,\sg(v)$, the surrogate
\begin{equation}
  \widetilde R=\sg\!\left(\|g\|_2^2\right)+s-\sg(s)
  \label{eq:r1-surrogate}
\end{equation}
has the value of the true penalty and its exact first derivative with respect
to $\phi$. The discriminator then minimizes
$\lambda_{\mathrm D}\big(\mathcal
L_{\mathrm{rel}}+\tfrac{\gamma}{2}(\widetilde R_1+\widetilde R_2)\big)$
with weights $\lambda_{\mathrm D}$ and $\gamma$. A single head forward on the
detached features $z$ and directions $v$ of the real and generated samples
produces both the relativistic logits and the directional terms, and one
ordinary backward pass updates the head (Figure~\ref{fig:exact-r1r2}). No step
differentiates through a backward kernel, and no finite difference or random
direction is involved.

\subsection{Streaming inference}
\label{sec:inference}

\begin{figure}[t]
  \centering
  \includegraphics[width=\linewidth]{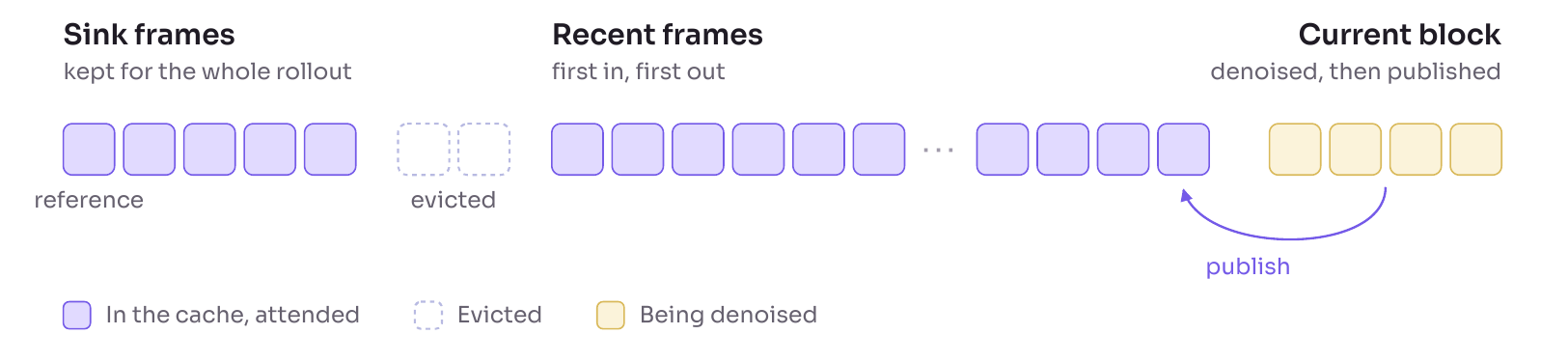}
  \caption{\captitle{The inference cache.} It retains a sink prefix, which begins
  with the reference, and recent frames, with configurable capacities; older
  frames are evicted. The current block attends to every cached frame and to
  the text. After denoising, the clean block joins the recent window.}
  \label{fig:kv-cache}
\end{figure}

The renderer prefills the reference and text, denoises incoming blocks, and
commits clean predictions into a bounded KV cache. The cache maintains a
permanent \emph{sink} prefix (including $x_0$) and a sliding window of recent
history (Figure~\ref{fig:kv-cache}). For rollouts exceeding the training
horizon, we apply top-aligned rotary position remapping: the active block is
clamped at the horizon boundary while recent history is shifted to precede it
and re-rotated, preserving relative temporal distances while geometry tracks
the true simulation timeline (Appendix~\ref{app:cache}).

\paragraph{Low-latency execution and deployment.}
Transformer execution over short blocks is primarily bounded by memory
bandwidth and kernel launch overhead. Rather than relying on
\texttt{torch.compile} \citep{ansel2024pytorch2}, which incurs minutes-long
cold-start compilation delays, we implement custom Triton kernels
\citep{tillet2019triton} to fuse elementwise operations around attention
(RMSNorm, RoPE rotation, and gated activations) without intermediate memory
round-trips. CUDA graph capture and replay further eliminate host-side launch
overhead for fixed-shape blocks. To bypass the heavy latent decoding bottleneck
during real-time serving, we optionally replace the standard VAE decoder with a
distilled tiny decoder \citep{bohan2025taehv}, which reconstructs pixel frames
in under 10\,ms (Table~\ref{tab:inference}). In interactive deployments,
the engine pushes conditions to a GPU worker, and decoded frames are streamed
to the agent or browser over WebRTC with bounded queuing latency
(Section~\ref{sec:inference-results}).

\section{Experiments}
\label{sec:evaluation}

We examine visual quality in long rollouts, replay accuracy,
and inference throughput. Performance measurements use one NVIDIA H100 GPU,
BF16 computation, $480\times832$ output, blocks of four latent frames, and
the four-step sampler unless stated otherwise.

\subsection{Self Forcing vs.\ Adversarial Forcing}
\label{sec:qualitative}
\label{sec:collapse}

Figure~\ref{fig:training-comparison} compares 30-second rollouts from
renderers trained with Self Forcing and with Adversarial Forcing on the same
condition trajectories. With Self Forcing, rollouts develop repetitive surface
patterns and lose scene detail as they grow longer. Adversarial Forcing
retains natural textures and fine detail throughout.

\newlength{\traininggap}\setlength{\traininggap}{1mm}
\newlength{\traininglabelw}\setlength{\traininglabelw}{17mm}
\newlength{\trainingw}
\setlength{\trainingw}{\dimexpr(\linewidth-\traininglabelw-4\traininggap)/4\relax}
\newcommand{\trainingimage}[2]{%
  \parbox[c]{\trainingw}{\includegraphics[width=\linewidth]{figures/images/training-#1-#2.jpg}}}
\newcommand{\trainingrow}[2]{%
  \noindent\makebox[\linewidth][l]{%
  \parbox[c]{\traininglabelw}{\raggedright\sffamily\footnotesize #2}\hspace{\traininggap}%
  \trainingimage{#1}{1}\hspace{\traininggap}%
  \trainingimage{#1}{2}\hspace{\traininggap}%
  \trainingimage{#1}{3}\hspace{\traininggap}%
  \trainingimage{#1}{4}}\par}
\newcommand{\trainingcase}[1]{%
  \trainingrow{#1-dmd}{Self Forcing}\vspace{1mm}%
  \trainingrow{#1-gan}{Adversarial Forcing}}
\newcommand{\trainingtimes}{%
  \noindent\makebox[\linewidth][l]{%
  \hspace{\traininglabelw}\hspace{\traininggap}%
  \makebox[\trainingw]{\sffamily\small 1\,s}\hspace{\traininggap}%
  \makebox[\trainingw]{\sffamily\small 10\,s}\hspace{\traininggap}%
  \makebox[\trainingw]{\sffamily\small 20\,s}\hspace{\traininggap}%
  \makebox[\trainingw]{\sffamily\small 30\,s}}\par\vspace{2mm}}
\begin{figure}[p]
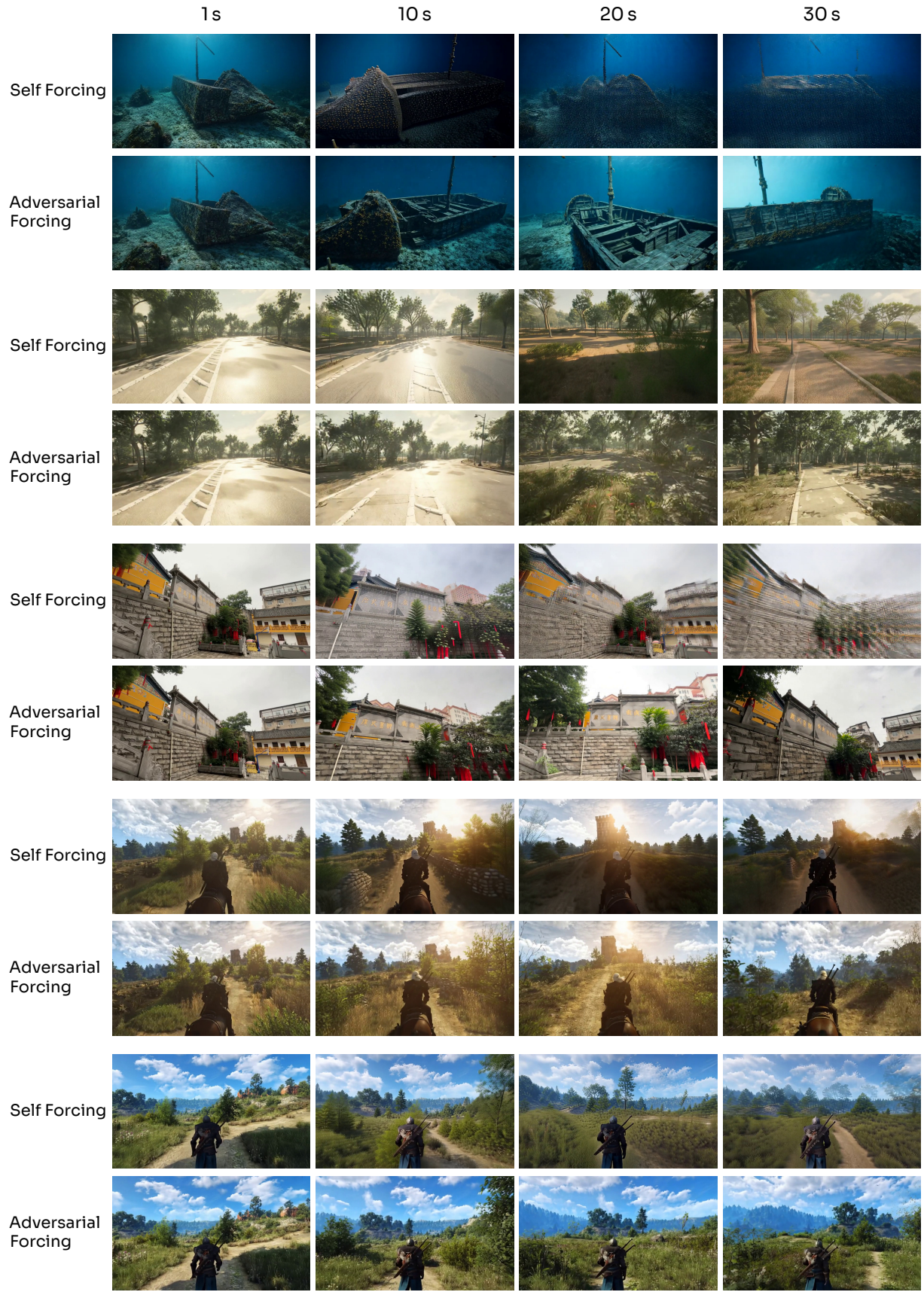

  \centering
  \setlength{\parskip}{0pt}
  \trainingtimes
  \trainingcase{underwater}
  \vspace{3mm}
  \trainingcase{road}
  \vspace{3mm}
  \trainingcase{temple}
  \vspace{3mm}
  \trainingcase{witcher-horse}
  \vspace{3mm}
  \trainingcase{witcher-field}
  \caption{\captitle{Qualitative comparison with Self Forcing.} Each pair shows
  the same 30-second condition trajectory at four moments.}
  \label{fig:training-comparison}
\end{figure}

\subsection{Replay accuracy}
\label{sec:replay-results}

Table~\ref{tab:replay} measures recomputation error and cost with only the
attention execution changed. Full-sequence FlexAttention replay differs from
the cached rollout by 3.99\% relative $L_2$ error; block-by-block SDPA replay
is bitwise identical. Forward time decreases by 5.4\%, while total peak memory
changes by 0.2\%.

\begin{table}[t]
  \centering
  \small
  \setlength{\tabcolsep}{6pt}
  \begin{tabularx}{\linewidth}{@{}Xccc@{}}
    \toprule
    \mirrosth{Measure} & \mirrosth{SGF-style (FlexAttention)} &
    \mirrosth{Ours (block-by-block SDPA)} & \mirrosth{Change} \\
    \midrule
    Replay error (relative $L_2$) & 3.99\% & 0 (bitwise) & --- \\
    \mirrosrowalt Forward time & 2.95\,s & 2.79\,s & $-5.4\%$ \\
    Peak allocated memory & 50.83\,GiB & 50.92\,GiB & $+0.2\%$ \\
    \bottomrule
  \end{tabularx}
  \caption{\captitle{Replay accuracy and cost.} Relative $L_2$ error between
  rollout and replay, measured on one H100 with 36 layers, $480\times832$ resolution,
  61 latent frames, and BF16. Times are medians of three warmed-up runs of the
  replay forward pass and exclude backward and optimizer updates. %
  }
  \label{tab:replay}
\end{table}

\subsection{Inference throughput}
\label{sec:inference-results}

With the hand-written kernels, CUDA graph capture, and the tiny decoder
(Section~\ref{sec:inference}), the renderer generates $480\times832$ video at
over \blockfps frames per second on one NVIDIA H100 GPU, including condition
encoding, four denoising steps, decoding, and host transfer.
Table~\ref{tab:inference} breaks one served block of sixteen frames into its
parts. The cache keeps five sink and 44 recent latent frames, one of the
history windows used in distillation (Appendix~\ref{app:cache}), so each
four-frame block attends to 49 frames of history. The block is measured at
steady state with the cache full: blocks 17 to 48 of a 48-block, 769-frame
rollout, averaged over two rollouts. Attention is dense and the tiny decoder
runs in FP16.

\begin{table}[t]
  \centering
  \small
  \setlength{\tabcolsep}{6pt}
  \begin{tabularx}{0.72\linewidth}{@{}Xr@{}}
    \toprule
    \mirrosth{One served block of 16 frames (one H100, BF16)} &
    \mirrosth{Time} \\
    \midrule
    Condition encoding & 10.6\,ms \\
    \mirrosrowalt Transformer: four denoising steps and cache publication & 414.4\,ms \\
    Tiny decoder and host transfer & 8.8\,ms \\
    \mirrosrowalt Total wall time & 438.8\,ms \\
    Throughput & 36.5 frames/s \\
    \bottomrule
  \end{tabularx}
  \caption{\captitle{Inference cost at steady state.} Measured with five sink
  and 44 recent latent frames in the cache and four current latent frames,
  using the hand-written kernels, CUDA graph capture, and the tiny decoder.
  Stage times are median GPU times; the total is the mean wall time per block
  and also covers work outside the three stages.}
  \label{tab:inference}
\end{table}

\section{Agents in Code Worlds}
\label{sec:agents}
\label{sec:applications}
\label{sec:environments}

Code worlds give agents many places to act, and a world that answers every
action is also a place to practice. This section presents environments built
on the shared renderer, revisits hide-and-seek with agents that perceive the
world only through rendered observations, describes the round-by-round
procedure through which they improve, and applies it to four further worlds.

\subsection{A library of environments}
\label{sec:library}

\begin{figure}[p]
  \centering
  \includegraphics[width=\linewidth]{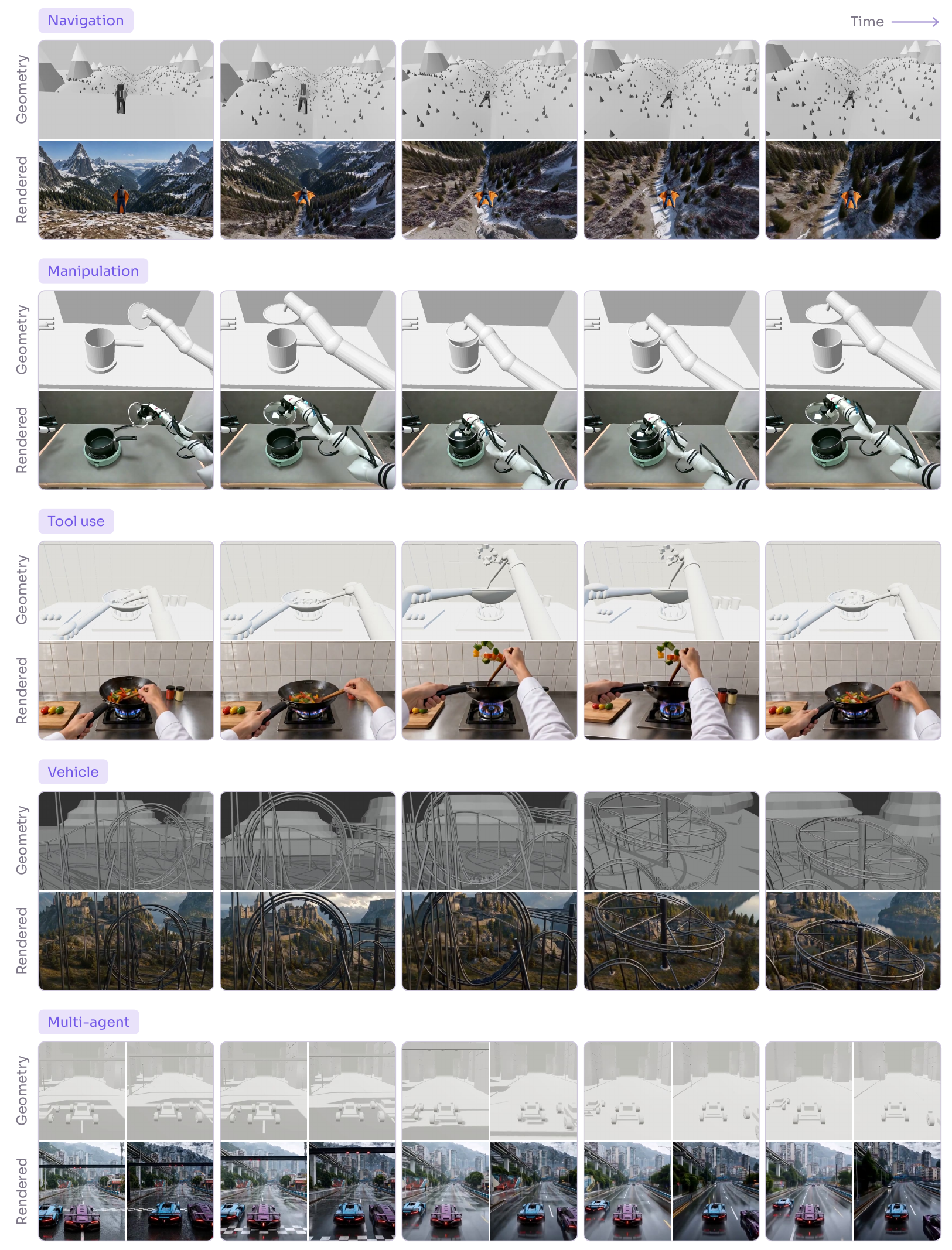}
  \caption{\captitle{Five code worlds, each at five moments of one rollout.} In
  every
  column, the untextured scene geometry (top) conditions the generated
  observation (bottom). Racing frames contain first-person views from two
  agents sharing one world state. The scene programs specify coarse geometry;
  the renderer supplies appearance from the initial image, text, and visual
  history.}
  \label{fig:qualitative}
\end{figure}

Different scene programs and compatible engines form code worlds with a
shared renderer (Figure~\ref{fig:qualitative}). The examples include
navigation, manipulation, tool use, and multi-agent interaction. Our
interactive deployment adds browser games whose rules live entirely in code,
including bowling, a penalty shootout, and a crate-vault puzzle; a player's
keyboard, mouse, or gamepad input advances the code world, and the rendered
stream is the only view of it. Because state evolution is explicit, it can
also be replayed and rendered again: a recorded rollout can be re-shot in a
different visual style, or from a different camera, without changing what
happened.

\subsection{Hide-and-seek, revisited}
\label{sec:hide-and-seek}

In OpenAI's hide-and-seek study \citep{baker2020emergent}, agents trained with
self-play and reinforcement learning developed strategies such as building
shelters, using ramps, and defending against those tools. The study reports
shelter construction after roughly 25 million episodes, followed by seeker
ramp use after another 75 million; the environment made a sequence of
increasingly sophisticated strategies possible through repeated competition.
Those agents observed object state. We revisit the setting with a pretrained
agent~\citep{openai2026gpt6astra} that can reason about what happened and
revise how it acts, and that perceives the world only through our renderer.

\paragraph{Setting.}
We use a sequential one-hider, one-seeker variant of the physics environment.
The hider prepares the scene and then hands control to the seeker. Each agent receives
first-person observations generated by the neural renderer from the
environment's structured conditions. It acts by submitting short Python
programs for movement and object interaction, and then observes the resulting
changes through the renderer. It never receives object coordinates, hidden
world state, or the opponent's private observations.

Each role keeps its own \emph{playbook}, organized as a library of short
\emph{skill files}, Markdown notes that may contain code, with one lesson per
file. The recorded run begins with empty playbooks and contains several rounds of
five games each, over sampled layouts and seeds. After each round, each role
reviews its own action programs and the visual evidence it was permitted to
see, identifies failures, and adds skill files to its playbook. In the next round, the agents receive the accumulated skills, interpret the current scene, and determine which prior experience is relevant. Both successful and unsuccessful attempts contribute to subsequent revisions.

\begin{figure}[t]
  \centering
  \includegraphics[width=\linewidth]{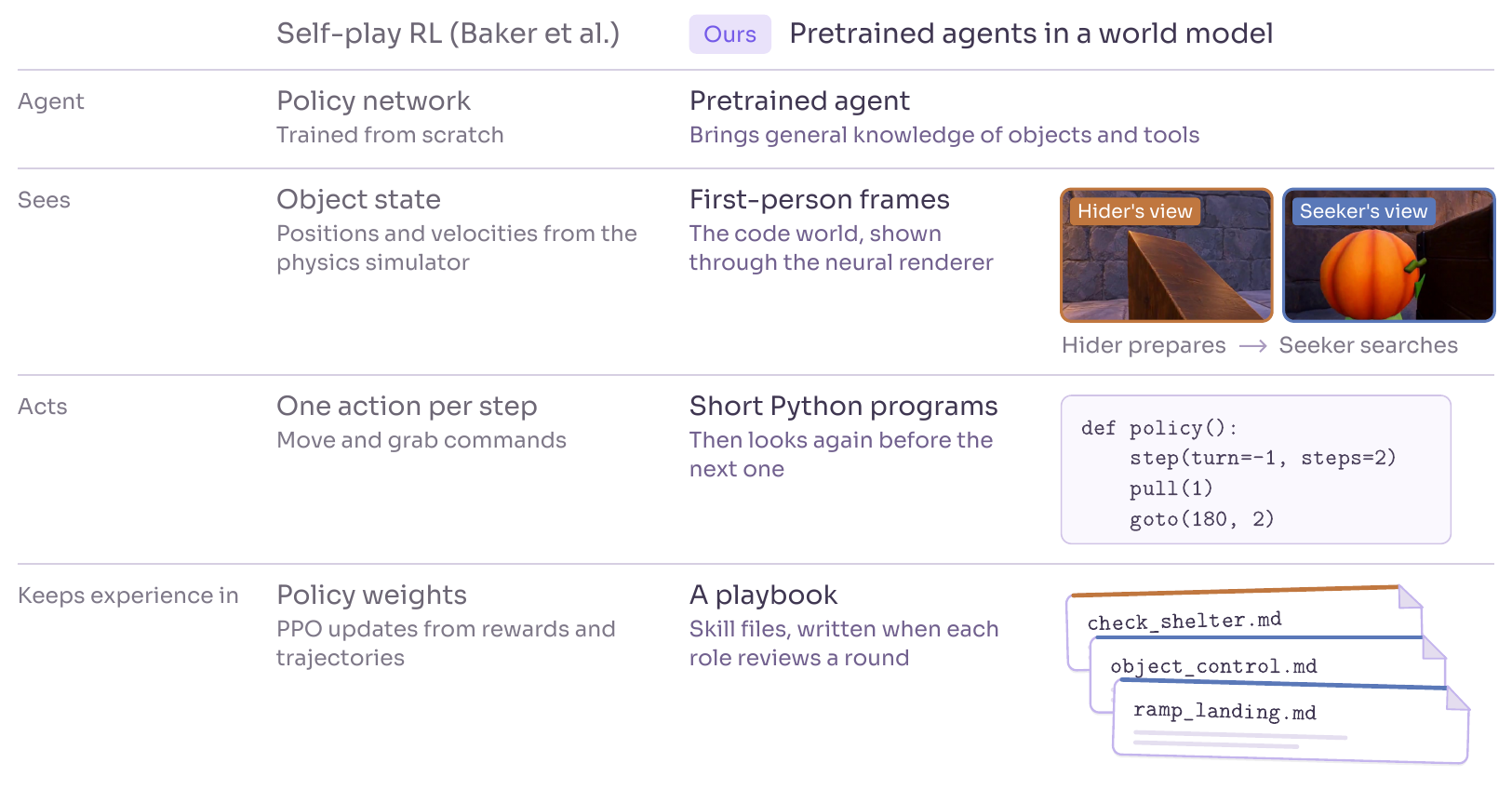}\\[3mm]
  \setlength{\tabcolsep}{1pt}%
  \begin{tabular}{@{}ccc@{}}
    \includegraphics[width=0.325\linewidth]{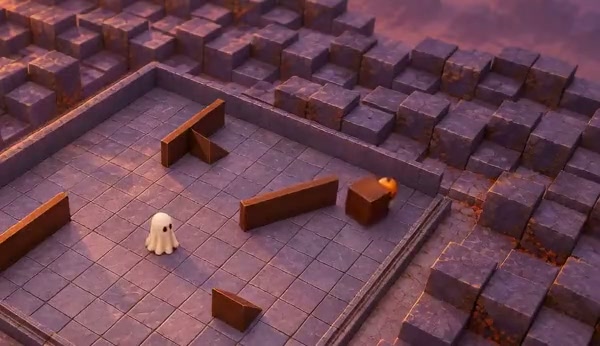} &
    \includegraphics[width=0.325\linewidth]{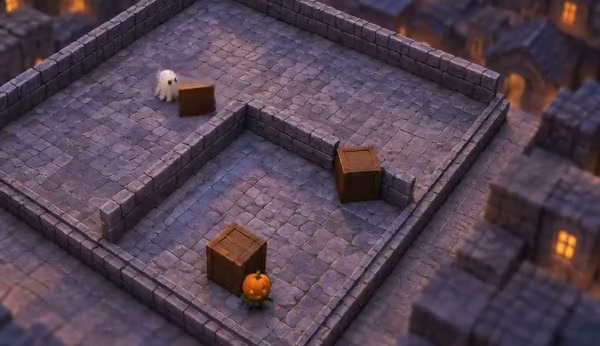} &
    \includegraphics[width=0.325\linewidth]{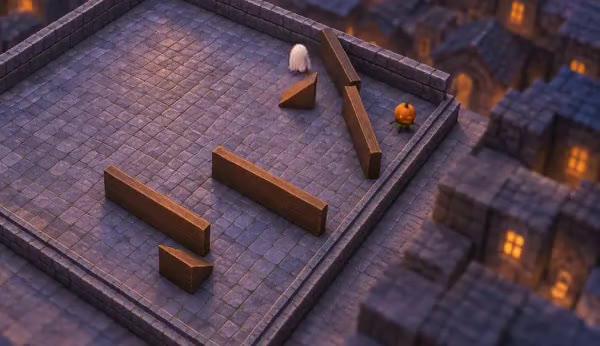} \\
    {\sffamily\scriptsize a hider moves a panel to rebuild cover} &
    {\sffamily\scriptsize a seeker carries a ramp over the wall} &
    {\sffamily\scriptsize a seeker moves the ramp closer and retries} \\
  \end{tabular}
  \caption{\captitle{Hide-and-seek in two settings.} Top: self-play
  reinforcement learning trains a policy network on object state and rewards.
  Our pretrained agents see first-person frames rendered by the neural
  renderer, act through short Python programs, and keep what they learn in a
  playbook of skill files. Bottom: three games from the run, seen from an overhead
  camera.}
  \label{fig:hns}
\end{figure}

\paragraph{Emergent behaviors.}
    Figure~\ref{fig:hns} highlights three characteristic behaviors that emerged during the interaction:
    a hider builds cover by moving a barrier panel to block an entrance;
    a seeker transports a ramp to an inner wall and climbs over it;
    and, after an initial leap falls short, a seeker repositions the ramp closer to the wall and tries again.
    These maneuvers demonstrate the agents' ability to acquire and refine multi-step physical tool-use strategies
  purely from neural-rendered first-person observations.

    These emergent strategies mirror the hallmark milestones of the 2019
    study, providing a reference point for physical tool use and counter-strategies
    in this domain (Table~\ref{tab:emerge}). There, shelter construction and ramp
    usage emerged after roughly 25 million and 100 million training episodes of
    self-play reinforcement learning, respectively.
    In our setting, the hiders used panels to build a shelter by round 4, and the seekers used a ramp to cross walls
  by round 10.

\begin{table}[t]
  \centering
  \small
  \setlength{\tabcolsep}{8pt}
  \begin{tabularx}{\linewidth}{@{}Xrr@{}}
    \toprule
    \mirrosth{Strategy} & \mirrosth{Self-play RL \citep{baker2020emergent}} &
    \mirrosth{Pretrained visual agents} \\
    & {\footnotesize training episodes} & {\footnotesize rounds played} \\
    \midrule
    Build shelters & ${\approx}\,25$ million & 4 \\
    \mirrosrowalt Use ramps to enter shelters & ${\approx}\,100$ million & 10 \\
    \bottomrule
  \end{tabularx}
  \caption{\captitle{Milestones of physical tool use in hide-and-seek under two distinct paradigms.}
  Self-play RL~\citep{baker2020emergent} trains tabula rasa policies over privileged object state across millions of episodes. In contrast, our study evaluates how pretrained agents, perceiving exclusively through the real-time neural renderer, ground general commonsense priors into closed-loop physical execution and adapt their strategies through written playbooks within a handful of games.}
  \label{tab:emerge}
\end{table}

Crucially, these numbers reflect two fundamentally different learning paradigms
rather than a direct sample-efficiency ratio. The 2019 agents learned physical
dynamics and competitive coordination from scratch with random weights,
operating directly on ground-truth coordinate state. In contrast, foundation model
agents already possess abstract knowledge of objects, tools, and geometry from
web-scale pretraining. Their primary challenge is not discovering concepts from
nothing, but grounding abstract knowledge into closed-loop sensorimotor action,
diagnosing spatial execution failures purely from synthesized visual observations
without state access, and refining tactical execution across rounds. This comparison
illustrates how combining a rich neural renderer with reflective playbooks enables
agents with general priors to bypass tabula rasa exploration and rapidly converge on
effective tool-use strategies in physical environments.

\subsection{Rounds of practice}
\label{sec:rounds}

\begin{figure}[t]
  \centering
  \includegraphics[width=\linewidth]{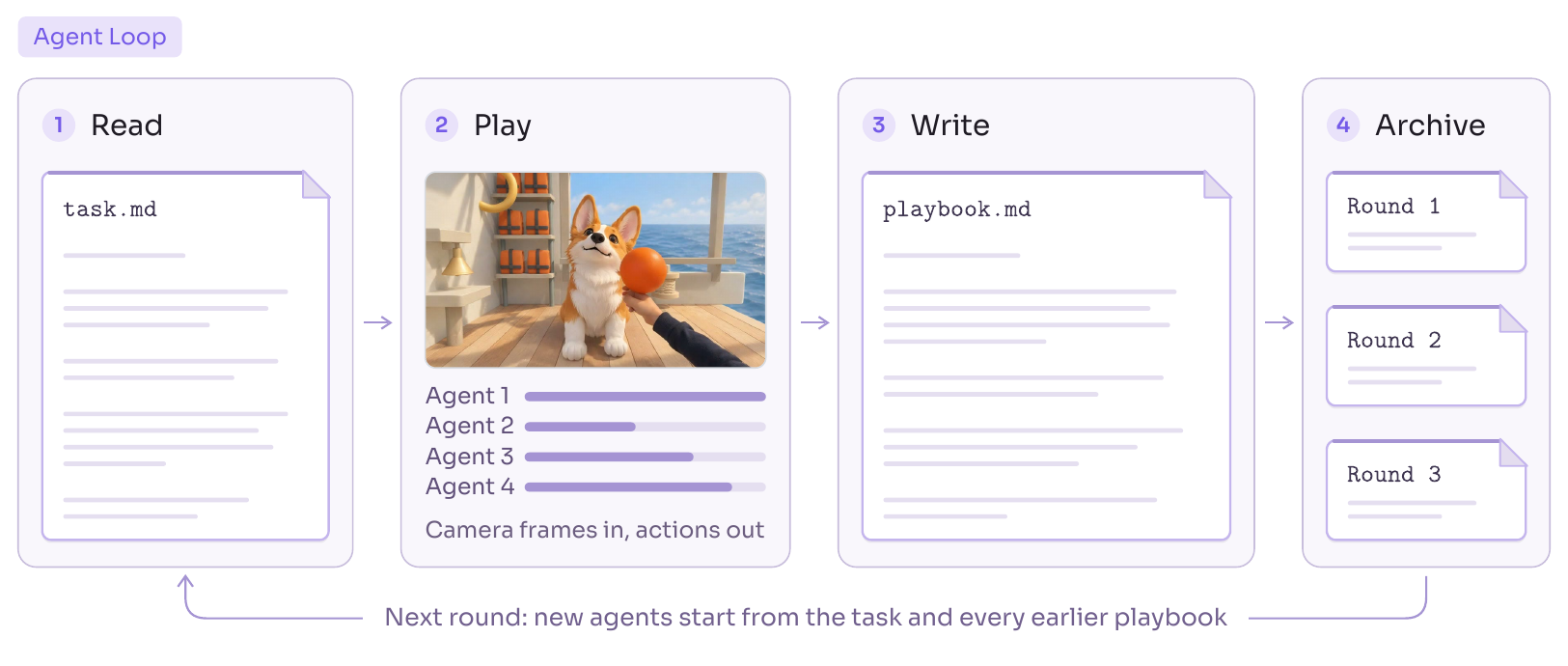}
  \caption{\captitle{One round of practice.} Agents read the task file,
  which states the goal, the actions, and the limits but no solution; play in
  parallel within a fixed budget, seeing only camera frames; and write a
  playbook. Playbooks are archived, and the next round's agents start in fresh
  conversations from the task file and every earlier playbook.}
  \label{fig:round-loop}
\end{figure}

Hide-and-seek is one instance of a general procedure
(Figure~\ref{fig:round-loop}). Practice runs in rounds. A round begins with a
\emph{task file} that states the goal, the available actions, and the limits,
but no solution. One or more agents then play in parallel, each within a fixed
budget of steps or simulated time. They see the world only through camera
frames synthesized by the neural renderer: no coordinates, no map, and no
score until the episode ends.
Afterwards each agent writes a playbook that records what it tried, what it
observed, what it still doubts, and what to test next. Playbooks are archived,
and the next round's agents, started in fresh conversations, receive the task
file and the playbooks of earlier rounds.

\subsection{Beyond hide-and-seek}
\label{sec:more-worlds}

We ran the same procedure in four more worlds
(Figure~\ref{fig:agent-worlds}); only the world and its task file change. In
the \emph{companion-dog} world an agent has
a 60-second session to keep a dog willingly engaged by offering a hand,
petting, and playing with a ball. On the \emph{one-lane bridge}, two cars, each
driven by its own agent from a windshield view, must swap ends of a bridge
that fits one car, as quickly as possible. In \emph{herding}, two dogs seeing
only from their own eye height guide four sheep into a pen and hold them there
for five seconds. In the \emph{quarry}, a wheel loader must push two rocks
onto staging pads, deliver one to a bunker behind a wall, and park, within 360
seconds. Each world has its own actions, time limit, and score, described to
the agent only in its task file (Appendix~\ref{app:tasks}), and each ran for four rounds. In all four
worlds, the agents act directly on first-person observations synthesized in
real time by the neural renderer, without access to internal simulation state
or raw engine geometry buffers. Figure~\ref{fig:agent-worlds} shows representative
frames from the recorded runs.

\newlength{\agentgap}\setlength{\agentgap}{1.2mm}
\newlength{\agentw}\setlength{\agentw}{\dimexpr(\linewidth-3\agentgap)/4\relax}
\newcommand{\agentframe}[2]{\includegraphics[width=\agentw]{figures/images/agents-#1-#2.jpg}}
\newcommand{\agentname}[1]{\makebox[\agentw][c]{\sffamily\footnotesize\bfseries\color{MirrosInk}#1}}
\newcommand{\agentnote}[1]{\makebox[\agentw][c]{\sffamily\scriptsize\color{MirrosMuted}#1}}
\begin{figure}[t]
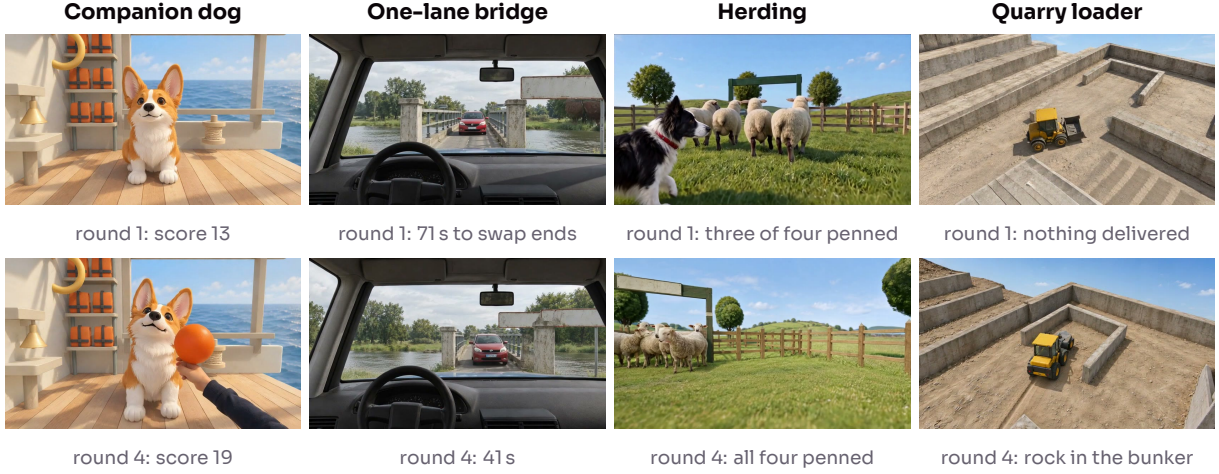

  \centering
  \noindent\makebox[\linewidth][l]{%
    \agentname{Companion dog}\hspace{\agentgap}\agentname{One-lane bridge}\hspace{\agentgap}%
    \agentname{Herding}\hspace{\agentgap}\agentname{Quarry loader}}\par\vspace{1mm}
  \noindent\makebox[\linewidth][l]{%
    \agentframe{corgi}{1}\hspace{\agentgap}\agentframe{bridge}{1}\hspace{\agentgap}%
    \agentframe{herding}{1}\hspace{\agentgap}\agentframe{loader}{1}}\par\vspace{0.4mm}
  \noindent\makebox[\linewidth][l]{%
    \agentnote{round 1: score 13}\hspace{\agentgap}\agentnote{round 1: 71\,s to swap ends}\hspace{\agentgap}%
    \agentnote{round 1: three of four penned}\hspace{\agentgap}\agentnote{round 1: nothing delivered}}\par\vspace{1.6mm}
  \noindent\makebox[\linewidth][l]{%
    \agentframe{corgi}{4}\hspace{\agentgap}\agentframe{bridge}{4}\hspace{\agentgap}%
    \agentframe{herding}{4}\hspace{\agentgap}\agentframe{loader}{4}}\par\vspace{0.4mm}
  \noindent\makebox[\linewidth][l]{%
    \agentnote{round 4: score 19}\hspace{\agentgap}\agentnote{round 4: 41\,s}\hspace{\agentgap}%
    \agentnote{round 4: all four penned}\hspace{\agentgap}\agentnote{round 4: rock in the bunker}}\par
  \caption{\captitle{The same loop in four further worlds.} A frame from
  round~1 (top) and round~4 (bottom) of each run, with the world's own
  measure. Table~\ref{tab:agent-worlds} lists every round.}
  \label{fig:agent-worlds}
\end{figure}

Table~\ref{tab:agent-worlds} lists the outcome of every round. In the
companion-dog world, the engagement score rose from 13 to 19. On the bridge, both cars arrived in every round,
and the time they needed fell from 71 to 41 seconds. The herding pair timed
out in round~1 with three of four sheep penned, and penned all four in every
later round. The loader cleared the rocks in round~1 and got no further,
scored nothing in round~2, and in round~4 cleared the rocks, delivered one, and
parked, in 329 of its 360 seconds.

\begin{table}[t]
  \centering
  \small
  \setlength{\tabcolsep}{6pt}
  \begin{tabularx}{\linewidth}{@{}lXrrrr@{}}
    \toprule
    \mirrosth{World} & \mirrosth{Measure} & \mirrosth{Round 1} & \mirrosth{2} &
    \mirrosth{3} & \mirrosth{4} \\
    \midrule
    Companion dog & engagement score & 13 & 14 & 19 & 19 \\
    \mirrosrowalt One-lane bridge & seconds until both cars arrive & 71 & 68 & 45 & 41 \\
    Herding & score out of 100 & 60 & 90.1 & 87.6 & 88.3 \\
    \mirrosrowalt & sheep penned, of four & 3 & 4 & 4 & 4 \\
    Quarry loader & score out of 100 & 30 & 0 & 30 & 90.9 \\
    \bottomrule
  \end{tabularx}
  \caption{\captitle{Outcome of every round in the four additional worlds.} Each row is
  the world's own terminal measure, one episode per round. For the bridge,
  lower is better.}
  \label{tab:agent-worlds}
\end{table}

\section{Related Work}
\label{sec:related-work}

\paragraph{Executable environments and code worlds.}
Simulated environments have long been the substrate of embodied learning.
Habitat~2.0 supports household rearrangement by simulated robots, while
ManiSkill2 provides a benchmark for diverse manipulation skills
\citep{szot2021habitat2,gu2023maniskill2}.
ProcTHOR generates interactive houses procedurally at scale
\citep{deitke2022procthor}, and the hide-and-seek study of
\citet{baker2020emergent} shows how a simple physics environment with
movable objects can support a sequence of increasingly sophisticated
strategies. Language models now automate environment construction: Holodeck
translates language into assets and spatial constraints, SceneCraft
synthesizes scene code with visual feedback, and LLMR generates and revises
interactive Unity experiences \citep{yang2023holodeck,hu2024scenecraft,delatorre2023llmr}.
Code2Worlds and SimWorld Studio extend this to dynamic scenes and to
environments with standard interfaces for embodied learning
\citep{zhang2026code2worlds,kang2026simworld}. A related line uses code to
recover dynamics rather than author them: WorldCoder infers transition and
reward functions from interaction, code world models translate the rules and
trajectories of a game into a program that a planner can search, and Code as
Worlds constructs and tests executable hypotheses from multimodal observations
\citep{tang2024worldcoder,lehrach2026codeworldmodels,wang2026codeasworlds}.
All of these share an explicit
representation that can be executed, inspected, and revised; their
observations, however, are rendered by conventional graphics pipelines and
inherit the fidelity of the available assets.

\paragraph{Explicit state behind a video model.}
Recent work also keeps the state of the world outside the video model and
uses the model to render it. The state may be learned jointly with the frames
\citep{lin2026stateplay,cai2026mass,meng2026marionette}, advanced by an engine
\citep{nvidia2026omnidreams,zhan2026magpie,chen2026worldnarrative,lin2026gamedirector},
or written as code by an agent
\citep{wang2026codeasworlds,chen2026codeworldmodel,huang2026programmable,yang2026oneira}.
These systems and ours share the separation of world evolution from
appearance. Our interface is depth or surface normals, which any simulator
with a 3D scene exports and which can be estimated from real videos; our
renderer is causal and runs in short blocks in real time, so new actions enter
throughout a rollout; and the state is a complete executable environment in
which pretrained agents act and improve over repeated rounds.

Rendering from coarse or intermediate representations also relates to neural
and generative rendering: from graphics buffers
\citep{cai2023generativerendering,liang2025diffusionrenderer,huang2026worldrenderer}
and from coarse 3D simulations of crowds
\citep{gomeznogales2026coarsetoreal}.

\paragraph{Interactive video world models.}
Video generation offers a learned route to interactive environments. Genie
learns latent actions from unlabeled video, and GameNGen simulates DOOM from
action--observation trajectories \citep{bruce2024genie,valevski2024gamengen}.
Later systems address control and memory over longer interactions: Matrix-Game
2.0 generates in real time from mouse and keyboard inputs, WorldPlay combines
action and camera conditioning with retrieved spatial context, Matrix-Game 3.0
adds error-aware rollout training and camera-aware retrieval, and SolarWM and
Astronex-World scale data and training for long-horizon, real-time interaction
\citep{he2025matrixgame2,sun2025worldplay,wang2026matrixgame3,huang2026solarwm,zhou2026astronex}.
Genie~3 adds world events triggered by text, Wonder improves camera control
and memory for real-time exploration, and ActWorld keeps the frames in which
an interaction happened so that changed objects stay changed
\citep{parkerholder2025genie3,xu2026wonder,xiong2026actworld}.
LingBot-World places pilot and director agents around a video world model to
steer actions and events \citep{gao2026lingbotworld2}. In these systems the
world is represented by generated history and learned memory. In ours, actions
first advance an executable environment, and the renderer is responsible only
for appearance; visual continuity is still learned, and generated frames need
not perfectly realize the supplied state.

\paragraph{Autoregressive video generation and distillation.}
CausVid adapts a bidirectional video diffusion model into a causal student
with asymmetric distribution matching \citep{yin2025causvid}. Self Forcing
trains an autoregressive student on its own rollouts rather than on
ground-truth prefixes \citep{huang2025selfforcing}, and LongLive extends this
to long sequences \citep{yang2025longlive}. Causal Forcing, Causal-rCM, and Mask
Forcing refine the initialization and rollout of self-forcing distillation
\citep{zhu2026causalforcing,zheng2026causalrcm,zhao2026maskforcing}. These
methods train the student with score distillation
\citep{yin2023dmd}; DMD2 adds a GAN loss on real data
\citep{yin2024dmd2}. Self Gradient Forcing \citep{zhuang2026selfgradient}
similarly introduces a second pass to differentiate history; our formulation
introduces exact replay, eliminating numerical divergence and matching the
rollout bitwise. Adversarial
post-training is also a route to few-step generation in its own right
\citep{lin2025apt,lin2025aapt}; its approximate R1 regularizer motivates our exact
alternative. With few sampling steps, decoding becomes a large share of the
remaining cost; tiny autoencoders distilled from a video VAE reduce it
\citep{bohan2025taehv}.

\paragraph{Agents that learn from written experience.}
Pretrained language models can improve at a task by keeping what they learn in
text. Reflexion has an agent reflect in words on a
failed attempt and carry the reflection into the next one
\citep{shinn2023reflexion}; ExpeL extracts reusable insights from a pool of
past trajectories \citep{zhao2024expel}; and Voyager grows a library of
executable skills while exploring Minecraft through a text interface to the
game's state \citep{wang2023voyager}. Our playbooks belong to this family.
What differs is the setting: the agents perceive the world through camera
frames, in hide-and-seek through the neural renderer alone, the task file
gives no solution, and playbooks pass between independent agents in fresh
conversations, so each lesson has to be stated well enough for another reader
to check it. Emergent strategy in hide-and-seek was first shown with self-play
reinforcement learning from random initialization \citep{baker2020emergent};
we ask what the same environment yields when the players start from a
pretrained model and a handful of games.

\paragraph{Gradient penalties and higher-order derivatives.}
R1 and R2 penalize the discriminator's input-gradient norm on real and
generated samples \citep{mescheder2018gans}; R3GAN combines them with a
relativistic loss into a stable modern baseline
\citep{jolicoeurmartineau2019relativistic,huang2024r3gan}. Computing their
parameter gradients requires differentiating an input gradient, which fused
attention kernels such as FlashAttention do not support
\citep{dao2022flashattention}. Forward-over-reverse products
\citep{pearlmutter1994hessian} and fused attention JVP kernels
\citep{zheng2025rcm} provide the ingredients we combine for an exact penalty
with a frozen discriminator backbone.

\section{Future Work}
\label{sec:limitations}
\label{sec:future-work}

\paragraph{More compact and complete conditions.}
Depth and surface normals provide a practical conditioning interface, but
dense geometry videos are redundant: once an object's shape is known, its
rigid motion is described by a few pose parameters, while a condition video
repeats the resulting surfaces across many pixels and frames. Spatial
downsampling reduces this cost but can remove thin structures, narrow gaps,
and small contact changes needed for precise control. Geometry alone also
leaves some of the world's state unspecified. A rotationally symmetric object
can spin without changing its depth or normals even as a painted marking
rotates, and material, color, and object identity are not determined by
geometry. Visual history can preserve these attributes, but may lose them
after long occlusions or revisits beyond the memory window. Future
conditioning interfaces could use more abstract and compact representations
of the state the code world already maintains, such as structured text
describing object attributes and interaction states, or high-dimensional
latent features. Structured text could, for example, specify the orientation
and angular velocity of a bullet spinning about its long axis even when its
depth and normal maps do not change. Aligning such representations with the
code world's state, and adapting pretrained video models to use them, remain
open problems.

\paragraph{Scaling environments.}
The interface also decides which worlds can be shown at all. A renderer that
receives only geometry cannot tell a spinning wheel from a still one, so
tasks that depend on such state are out of reach today. A richer interface
widens the range of possible worlds, but someone still has to write them. We
built the worlds in this report one at a time. Since a world is a program,
the next step is to let coding agents write and revise
worlds~\citep{wang2026codeasworlds,chen2026codeworldmodel}, with the shared
renderer supplying appearance. Generating a world is the easy part. The
harder question is whether it deserves an agent's time: whether the task can
be solved from what the agent sees, whether a careless strategy fails, and
whether there is something to learn that carries into the next round. We want
these checks to run automatically, before any agent practices in a new world.

\paragraph{Scaling experience.}
More worlds change how experience has to be kept. Here each playbook belongs
to one world and is short enough to read in full. Across many worlds, agents
must decide which lessons to keep, merge, or drop, and find the few that
apply to the scene in front of them. Code worlds help: a world can be reset
and replayed exactly, so a lesson can be tested before it is passed on.
Lessons that hold in several worlds, such as how to confirm that a grasp held
or which visual cues mislead, become general knowledge about acting in
physical scenes, and unseen worlds can measure it. The two grow together:
where agents fail tells us which worlds to write next, and each new world
tests what they wrote down before.

\section{Conclusion}
\label{sec:conclusion}

What an agent can learn is bounded by the environment it practices in.
AgentGarten builds environments that are both faithful and realistic by
dividing the work: a simulator or game engine maintains persistent state and
executes the rules of each world, and a shared neural renderer, trained with
Adversarial Forcing, turns the geometry it exports into observations in real
time. An agent can therefore act, see the consequence, and act again, in a
world whose state it can rely on. In these environments, pretrained agents
improve from their own experience: they play, review each round, and write
playbooks that later agents inherit and build upon. In hide-and-seek, seen
only through rendered observations, shelters appeared by round~4 and ramp
crossings by round~10, and the same procedure improved outcomes in four
further worlds. Because a new world is written as code and rendered through
the same interface, environments can grow in number and difficulty alongside
the agents that practice in them. We expect agents that keep evolving
through interaction to come from that loop.

\section*{Authors}
\phantomsection
\addcontentsline{toc}{section}{Authors}
Jiawei Chi,
Shangchen Miao,
Zhiyuan Shi,
Kailu Wu,
Hanyang Wang,
Weiliang Chen,
Qiyu Dai,
Chengyang Li,
Jinshan Ren,
Jun Gao,
Mingsheng Long,
Yueqi Duan,
Jiangran Lyu,
Jialong Wu\corrauthor,
Fangfu Liu\corrauthor
\vspace{\baselineskip}

\section*{Affiliations}
\phantomsection
\addcontentsline{toc}{section}{Affiliations}
MirroS, Tsinghua University, Peking University, Shanghai Jiao Tong University
\vspace{\baselineskip}

\begingroup
\renewcommand{\thefootnote}{\fnsymbol{footnote}}
\footnotetext[2]{Project lead.}
\endgroup

\clearpage
\printbibliography[heading=bibintoc]

\clearpage
\appendix

\section{Implementation Details}
\label{app:implementation}

\subsection{Backbone, data, and training stages}
\label{app:stages}

The renderer uses the video stream of Cosmos~3-Nano~\citep{nvidia2026cosmos3}:
36 Transformer layers with hidden width 4096, 32 query heads, and 8
key--value heads. The text stream and the Wan video autoencoder
\citep{wanteam2025wan} stay frozen; text encoding and video encoding are
computed online. The autoencoder compresses time by four and space by sixteen,
and a $2\times2$ patch embedding produces \rgbtokens tokens per latent frame at
$480\times832$. The model's temporal grid is treated as 16 frames per second.
Five-second training windows contain 81 frames (a reference and 20 latent
frames, five target blocks); fifteen-second windows contain 241 frames (a
reference and 60 latent frames, fifteen target blocks).

Training clips pair RGB video with a text description and with depth and
normals, estimated as described in Section~\ref{sec:conditions}. The mixture
contains scene captures from DL3DV-10K~\citep{ling2024dl3dv}, driving videos
from OpenDV~\citep{yang2024genad}, and internet videos of gameplay,
navigation, and robot interaction, supplemented by a small set of rendered
synthetic scenes, whose conditions are exported by the renderer instead of
estimated.

Table~\ref{tab:stages} summarizes the training stages. All stages run on 32
NVIDIA H100 GPUs with fully sharded data parallelism, one clip per GPU, and two
gradient-accumulation steps, for 64 clips per optimizer update. Computation
uses BF16 with FP32 gradient reductions; the discriminator head uses FP32
parameters. All optimizers are AdamW with zero weight decay. Adaptation stages
use 100-update linear warmup from 10\% of the listed
rate; distillation uses constant rates without warmup.

\begin{table}[!ht]
  \centering
  \small
  \setlength{\tabcolsep}{4pt}
  \begin{tabularx}{\linewidth}{@{}l>{\raggedright\arraybackslash}Xlcc@{}}
    \toprule
    \mirrosth{Stage} & \mirrosth{Updated parameters} & \mirrosth{Initialization} &
    \mirrosth{LR} & \mirrosth{Adam $(\beta_1,\beta_2)$} \\
    \midrule
    1. Geometry conditioning & self-attention projections and q/k norms; condition embedding &
      Cosmos 3-Nano & $3\times10^{-5}$ & $(0.9,0.99)$ \\
    \mirrosrowalt 2. Causal adaptation & same as stage 1 & stage 1 & $3\times10^{-5}$ & $(0.9,0.99)$ \\
    3. Distillation: student & complete video stream & stage 2 & $2\times10^{-6}$ & $(0,0.999)$ \\
    \mirrosrowalt \quad fake score & complete video stream & stage 1 & $4\times10^{-7}$ & $(0,0.999)$ \\
    \quad discriminator & head only; backbone is the frozen teacher & fresh & $2\times10^{-7}$ & $(0,0.999)$ \\
    \bottomrule
  \end{tabularx}
  \caption{Training stages. Stages 1 and 2 train first on five-second and then
  on fifteen-second windows; each distillation run uses the teacher and fake
  score of the matching window length. The teacher is the stage-1 model and
  stays frozen. The text stream and the video autoencoder are frozen
  throughout.}
  \label{tab:stages}
\end{table}

\subsection{Geometry preprocessing}
\label{app:geometry}

Following Vision Banana~\citep{visionbanana2026}, valid depth $d>0$ is mapped to
\begin{equation}
  u(d)=1-\left(1+\frac{\alpha d}{10}\right)^{-2}.
\end{equation}
The scale $\alpha$ is shared across the clip. Metric depth uses $\alpha=1$;
for depth with an unknown scale, the reader estimates the median $m$ of valid
depth samples across the clip and sets $\alpha=10(\sqrt{2}-1)/m$, placing that
median at $u=0.5$. Training can jitter this scale for non-metric clips. The
coordinate $u\in[0,1)$ traverses seven equal segments of the RGB cube:
black, red, yellow, green, cyan, blue, magenta, and white, with linear
interpolation within each segment. Invalid depths are mapped to black.

Normals are kept in $[-1,1]$ at the autoencoder input, corresponding to the
display encoding $(n+1)/2$ in RGB;
estimated normals are canonicalized by negating the $x$ component of the
NormalCrafter prediction, and rendered normals are converted to the same
camera-frame orientation. When a modality is dropped, its video is set to the
value that becomes $-1$ (black) at the autoencoder input.

The condition is average-pooled by four along each spatial axis and placed on
the condition canvas with a piecewise-bilinear stretch. At $480\times832$
output resolution this gives \geotokens condition tokens per latent frame,
compared with \rgbtokens RGB tokens.
For spatial rotary positions, a condition-grid coordinate $u$ maps to
$(u+\tfrac12)(N_{\mathrm{rgb}}/N_{\mathrm{geo}})-\tfrac12$ along each axis,
where $N_{\mathrm{rgb}}$ and $N_{\mathrm{geo}}$ are the respective grid sizes.
Texture suppression
(probability $0.5$) downsamples by a factor drawn from $[1.5,3]$. Spatial augmentation (probability $0.5$) applies a smooth warp anchored
to the first frame, with scale at most $1.15$, that decays over the clip. A
per-sample choice keeps depth or normals, and with probability $0.1$ both are
dropped. After encoding, Gaussian latent noise with standard deviation $0.4$
is added to pooled conditions.

\subsection{Causal blocks and history perturbation}
\label{app:causal}

The reference is a standalone clean prefix. It is never assigned a diffusion
timestep, never included in the loss, and never republished after the cache is
initialized. Each subsequent block contains four latent frames. For every
target block, the clean and noisy copies carry separate condition tokens and
follow Eq.~\eqref{eq:ar-attention-mask}; flow times are sampled independently
for each target block, and the clean copy is evaluated at the clean-context
timestep. Each visible history frame independently receives one of four equally
likely perturbations: contrast and exposure scaling about its channel mean with
a factor from $[0.3,1.7]$, Gaussian noise mixing with strength from $[0,1/3]$,
bilinear down- and upsampling with scale from $[0.9,1]$, or no change. The
reference frame is never perturbed.

\subsection{Distillation}
\label{app:distillation}

For the four-step renderer, the sampling trajectory has endpoints in
normalized flow time
\begin{equation}
  \mathcal{E}=(1600/1601,15/16,5/6,5/8,0).
\label{eq:distillation-schedules}
\end{equation}
All blocks in a rollout use the same trajectory, and the replay recomputes its
final step. The teacher uses classifier-free guidance scale
6, score times follow a shifted uniform distribution with shift 5, and the
fake-score model takes five updates per student update.
Clean estimates are obtained from velocity predictions as
$f(y_\tau,\tau)=y_\tau-\tau v(y_\tau,\tau)$, and the normalization in
Eq.~\eqref{eq:dmd} is
\begin{equation}
  a=\max\!\left\{
    \operatorname{mean}\!\left(
      |\widetilde z_\theta-f_{\mathrm{T}}(y_\tau,\tau\mid\mathcal{C})|
    \right),10^{-5}\right\},
\end{equation}
with the mean over all latent elements of each video. The score estimates,
normalization, and regression target all use the second-pass prediction of
Eq.~\eqref{eq:sgf-prediction}; the recorded first-pass output supplies context
but is not substituted for the prediction. The DMD loss carries a factor
$0.5$, included in the $1/(2N)$ coefficient of Eq.~\eqref{eq:dmd}.

\paragraph{Exact replay.}
The replay recomputes the clean-history keys and values within the
differentiable graph; it never reads detached cache values from the rollout.
Attention is computed one block at a time, as in the cached rollout: once for
the block's noisy queries and once for its clean publication, each call with
the same visible context, key--value order, and tensor shapes. Projections and feed-forward layers are grouped
the same way: the reference first, then, for each block, its condition tokens
followed by its RGB tokens. Grouping matters because a projection applied to a
whole packed sequence can round differently from the same projection applied
block by block, and such differences amplify through a trained network. The
grouping loop runs outside the compiled region while the per-group tensor
operations stay compiled, so changes in history length or caption length do not
recompile an unrolled layer.

\paragraph{Adversarial term.}
The discriminator reads the frozen teacher's features after layers 11, 23, and
35. Each branch aggregates its layer's tokens with a learned query followed by
a residual multilayer perceptron, and the concatenated branch outputs give one
logit. With the critic parameters fixed, let
$h=\partial\mathcal{L}_{\mathrm{G}}/\partial\widetilde z_\theta$ be the
generator-side gradient; it enters the student graph through the linear
surrogate $\langle\widetilde z_\theta,\sg(h)\rangle$, so that the adversarial
and DMD objectives both act through the replay without retaining the rollout
graph.

\section{Exact R1/R2}
\label{app:r1r2}

\paragraph{Why freezing the backbone matters.}
Section~\ref{sec:gan} writes $z=B(x)$, $A=J_B(x)$, $u=\nabla_z h_\phi(z)$, and
obtains $g=A^\top u$ by a VJP and $v=Ag$ by a JVP. Because the backbone is
frozen, $z$ and $A$ are independent of $\phi$, and differentiating $R=g^\top g$
gives Eq.~\eqref{eq:r1-identity}; the factor of two accounts for the two copies
of $g$. The direction $v$ is recomputed at the current parameters on every
update and then detached. A nonlinear backbone is fully compatible with the
identity: its Jacobian depends on $x$ but not on $\phi$. Updating the backbone
as well would introduce derivatives of its features and Jacobian that the
detached construction does not supply. Freezing alone does not remove the
head's own second-order derivatives; the explicit head JVP below changes how
they are computed.

\paragraph{Why JVPs are simpler to implement.}
Backward-over-backward differentiates the computation that produced the input
gradient. For a fused attention kernel this requires a differentiable backward
or a separate implementation of its derivatives, with correct gradient paths
through saved intermediates and incoming gradients. A JVP instead follows the
forward computation, carrying each activation together with one tangent
\citep{pearlmutter1994hessian}. Each operator needs only a local directional
rule: a linear layer propagates $\dot y=W\dot x$, and a residual addition adds
the two tangents. These rules compose in the order of the original network and
can be checked operator by operator. For attention $Y=PV$ with
$P=\operatorname{softmax}(S)$ and $S=QK^\top/\sqrt d$,
\begin{equation}
\begin{aligned}
  \dot S&=(\dot QK^\top+Q\dot K^\top)/\sqrt d,\\
  \dot P&=P\odot\Big(\dot S-\textstyle\sum_{\mathrm{keys}}P\odot\dot S\Big),\\
  \dot Y&=\dot P V+P\dot V,
\end{aligned}
\end{equation}
where the sum runs over keys within each query row. The frozen backbone needs
only the resulting tangent values, so its replay runs without a gradient graph,
using a fused attention-JVP kernel anchored at the activations recorded by the
original forward pass; the fixed history, conditions, and text have zero
tangent. The trainable head, whose cross-attention branches each have a single
query token, expresses the same rules with ordinary matrix products, softmax,
and elementwise operations; its attention maps grow only linearly with the
number of feature tokens, so the dense differentiable implementation is
practical. An ordinary backward through the directional score $s$ then supplies
the required mixed derivative. Mathematically this is still a second-order
derivative of the head; computationally, no FlashAttention backward kernel is
ever differentiated. rCM's FlashAttention JVP kernel \citep{zheng2025rcm}
computes attention and its tangent in one fused forward pass; our backbone
needs only such values, whereas our head also needs gradients through the JVP
with respect to its parameters, which the explicit head JVP preserves.

\paragraph{The surrogate.}
The surrogate $\widetilde R$ in Eq.~\eqref{eq:r1-surrogate} reports the full
squared input-gradient norm while routing the head gradient through $s$. It
matches the penalty and its first derivative with respect to $\phi$ at the
current update; it is not a replacement graph for arbitrary higher-order
derivatives. R1 and R2 apply the construction to real and generated inputs,
with generated inputs detached during the discriminator update, and the
discriminator minimizes $\lambda_{\mathrm D}(\mathcal L_{\mathrm{rel}}+
\tfrac{\gamma}{2}(\widetilde R_1+\widetilde R_2))$. In code, the real and
generated preparations run sequentially, so only one backbone graph exists at
a time, and one head forward returns the pair logits and both directional
scores.

\section{Streaming Cache}
\label{app:cache}

\paragraph{Cache policy.}
Every query attends to the text, the current block, and all cached history.
The cache is bounded by a number of sink latent frames, which include the
reference, and a number of most recent latent frames. During distillation,
each rollout draws its history window from three such settings: no sink and
48 recent frames, one sink and 48 recent frames, or five sink and 44 recent
frames. The served configuration retains five sink and 44 recent frames, the
last of these settings; Table~\ref{tab:inference} reports its throughput.
Eviction removes a frame's RGB and condition keys
and values together.

\paragraph{Top-aligned rotary positions.}
Let $H$ be the training horizon in latent frames, $b$ the block length, and
$s$ the next block's start on the simulation timeline. Its rotary start is
$\min(s,H-b)$, so positions remain within the training range. Recent history
occupies consecutive positions immediately before that block, while the sink
prefix keeps its original positions. We re-rotate retained history keys from
their previous to their new coordinates using the backbone's native rotary
templates, including its frame-rate and modality offsets. Values are moved
without rotation.
Condition contents are indexed by the simulation timeline, independently of
this position remapping. The implementation uses $H=61$ and $b=4$ for the
fifteen-second model.

\section{Task Files}
\label{app:tasks}

A task file is all an agent is told about its world before it plays: the
goal, the available actions, and the limits (Section~\ref{sec:applications}).
The listings below reproduce the task files of the four additional worlds
verbatim, as given to one agent in round~1; long lines are wrapped. In the
bridge and herding worlds the second agent receives the same file for its own
color.

\lstdefinestyle{taskfile}{
  basicstyle=\ttfamily\footnotesize\color{MirrosInk},
  columns=fullflexible,
  keepspaces=true,
  breaklines=true,
  breakindent=1.5em,
  aboveskip=0pt,
  belowskip=0pt,
  literate={—}{{---}}1,
  moredelim=[l][\bfseries\color{MirrosPurpleSoft}]{\#},
}
\newtcbinputlisting{\taskfile}[2]{%
  listing file={#2}, listing only, listing options={style=taskfile},
  enhanced, breakable,
  colback=MirrosCardBg!40, colframe=MirrosLine, boxrule=0.5pt, arc=3pt,
  left=9pt, right=9pt, top=6pt, bottom=6pt,
  title={{\ttfamily task.md}\hfill{\sffamily\bfseries #1}},
  fonttitle=\footnotesize, coltitle=MirrosInk, colbacktitle=MirrosCardBg,
  toptitle=3pt, bottomtitle=3pt, titlerule=0.5pt,
  before skip=0.9\baselineskip, after skip=1.1\baselineskip,
}

\taskfile{Companion dog}{sections/tasks/corgi.md}

\taskfile{One-lane bridge · blue car}{sections/tasks/bridge.md}

\taskfile{Herding · blue dog}{sections/tasks/herding.md}

\taskfile{Quarry loader}{sections/tasks/loader.md}

\end{document}